\documentclass{article}

\usepackage[preprint, nonatbib]{neurips_2020}

\usepackage[utf8]{inputenc} % allow utf-8 input
\usepackage[T1]{fontenc}    % use 8-bit T1 fonts
\usepackage{graphicx}
\usepackage{booktabs} % for professional tables

\usepackage{algorithm}
\usepackage{algorithmic}
\usepackage[numbers]{natbib}
\usepackage{eso-pic} % used by \AddToShipoutPicture 
\usepackage{forloop}

\usepackage{amsfonts}       % blackboard math symbols
\usepackage{nicefrac}       % compact symbols for 1/2, etc.
\usepackage{microtype}      % microtypography

\usepackage[hidelinks]{hyperref}

\usepackage{url}
\usepackage{amsmath}
\usepackage{graphicx}
\usepackage{subcaption}
\usepackage{wrapfig}
\usepackage{adjustbox}
\usepackage{tabularx}
\usepackage{bbm}
\usepackage{listings}
\PassOptionsToPackage{hyphens}{url}

\usepackage{autobreak}
\allowdisplaybreaks

\usepackage{amsmath,amsfonts,bm}

\def\eqref#1{eq.~\ref{#1}}
\def\Eqref#1{Eq.~\ref{#1}}
\def\1{\bm{1}}

\DeclareMathAlphabet{\mathsfit}{\encodingdefault}{\sfdefault}{m}{sl}
\SetMathAlphabet{\mathsfit}{bold}{\encodingdefault}{\sfdefault}{bx}{n}

\DeclareMathOperator*{\E}{\mathbb{E}}

\DeclareMathOperator*{\argmin}{arg\,min}

\newsavebox{\bigimage}

\newcommand{\prob}[1]{\ensuremath{\text{P}\left(#1\right)}}

\def \bx{\boldsymbol{x}}
\def \by{\boldsymbol{y}}
\def \bz{\boldsymbol{z}}

\def \bpsi{\boldsymbol{\psi}}

\makeatletter
\newcommand\footnoteref[1]{\protected@xdef\@thefnmark{\ref{#1}}\@footnotemark}
\makeatother

\title{Black-Box Optimization with Local Generative Surrogates}
\author{%
  Sergey Shirobokov\thanks{Equal contribution} \\
  Department of Physics\\ 
  Imperial College London\\
  United Kingdom\\
  \texttt{s.shirobokov17@imperial.ac.uk} \\
   \And
   Vladislav Belavin$^{*}$ \\
   National Research University \\
   Higher School of Economics \\
   Moscow, Russia \\
   \texttt{vbelavin@hse.ru} \\
   \AND
   Michael Kagan \\
   SLAC National Accelerator Laboratory\\
   Menlo Park, CA \\
   United States \\
   \And
   Andrey Ustyuzhanin \\
   National Research University \\
   Higher School of Economics \\
   Moscow, Russia \\
   \And
   Atılım Güneş Baydin \\
   Department of Engineering Science\\
   University of Oxford\\
   United Kingdom \\
}

\begin{document}

\maketitle

\begin{abstract}
We propose a novel method for gradient-based optimization of black-box simulators using differentiable local surrogate models. In fields such as physics and engineering, many processes are modeled with non-differentiable simulators with intractable likelihoods. Optimization of these forward models is particularly challenging, especially when the simulator is stochastic. To address such cases, we introduce the use of deep generative models to iteratively approximate the simulator in local neighborhoods of the parameter space. We demonstrate that these local surrogates can be used to approximate the gradient of the simulator, and thus enable gradient-based optimization of simulator parameters. In cases where the dependence of the simulator on the parameter space is constrained to a low dimensional submanifold, we observe that our method attains minima faster than baseline methods, including Bayesian optimization, numerical optimization, and approaches using score function gradient estimators.
\end{abstract}

\section{Introduction}
Computer simulation is a powerful method that allows for the modeling of complex real-world systems and the estimation of a system's parameters given conditions and constraints. Simulators drive research in many fields of engineering and science \citep{cranmer2019frontier} and are also used for the generation of synthetic labeled data for various tasks in machine learning \citep{ruiz2018learning,richter2016playing,ros2016synthia}. A common challenge is to find optimal parameters of a system in terms of a given objective function, e.g., to optimize a real-world system's design or efficiency using the simulator as a proxy, or to calibrate a simulator to generate data that match a real-data distribution. A typical simulator optimization problem can be defined as finding $\bpsi^{*} = \argmin_{\bpsi} \sum_{\bx} \mathcal{R}(F(\bx, \bpsi))$, where $\mathcal{R}$ is an objective we would like to minimize and $F$ is a simulator that we take as a black box with parameters $\bpsi \in \mathbb{R}^{n}$ and inputs $\bx \in \mathbb{R}^{d}$.

\begin{wrapfigure}{R}{0.5\textwidth}
\vspace{- \baselineskip}
    \includegraphics[width=\linewidth]{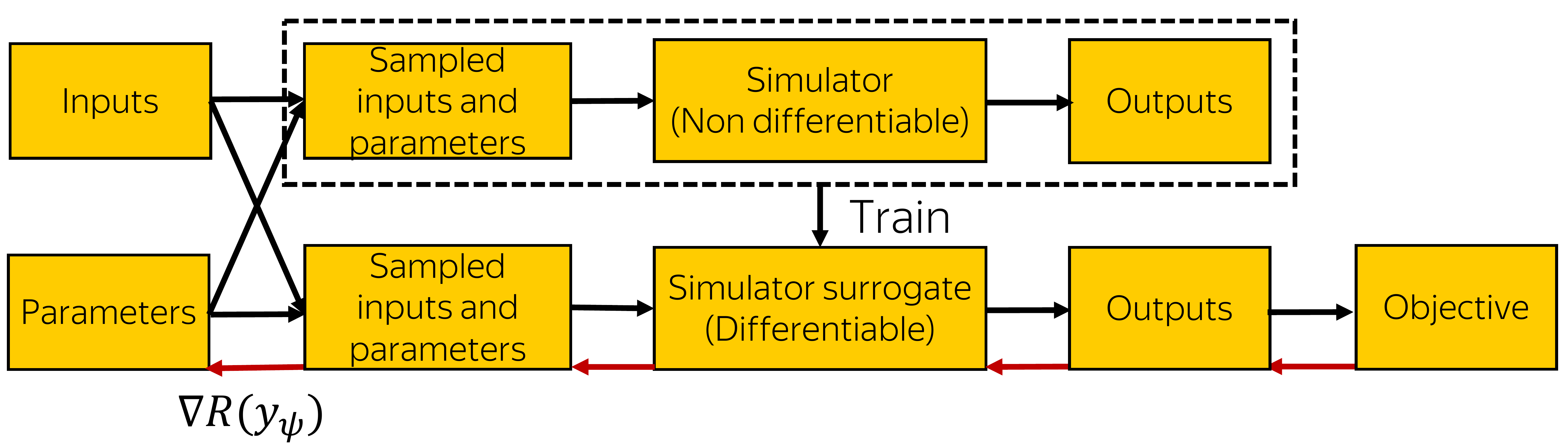}
    % \centering
    % \includegraphics[trim=9 79 14 21,clip,width=0.9\linewidth]{figures/pdf_plots/gra.pdf}
    \caption{Simulation and surrogate training. \emph{Black:} forward propagation. \emph{Red:} error backpropagation.}
    % For further details refer to Algorithm~\ref{alg:euclid_local}.}
	\label{Fig:algo_graph}
\end{wrapfigure}%

In this work, we focus on cases where the simulator and its inputs are stochastic, so that $\by = F(\bx, \bpsi)$ is a random variable $\by \sim p(\by | \bx; \bpsi)$, the inputs are $\bx \sim q(\bx)$, and the objective is expressed as the expectation $\E_{p(\by| \bx; \bpsi)}[\mathcal{R}(\by)]$. 
Note that the choice of modeling the simulator inputs $\bx$ as random reflects the situation common in scientific simulation settings, and our methods are equally applicable for the case without stochastic inputs such that $\by \sim p(\by; \bpsi)$.

In many settings the cost of running the simulator is high, and thus we aim to minimize the number of simulator calls needed for optimization. Such stochastic simulator optimization occurs in an array of scientific and engineering domains, especially in cases of simulation-based optimization relying on Monte Carlo techniques. Examples include optimization for particle scattering simulators \citep{sjostrand2008brief}, radiation transport simulators~\citep{Amadio_2017}, and molecular dynamics simulation~\citep{abraham2011optimization}. In some domains, this is also referred to as design optimization~\citep{papalambros_wilde_2000}.
 
Several methods exist for such optimization, depending on the availability of gradients for the objective function~\citep{design_opt_overview}.  While gradient-based optimization has been demonstrated to work well for differentiable objective functions~\citep{ChainQueen,chang2016compositional,BelbutePeres2018EndtoEndDP,Degrave2017ADP}, non-differentiable objective functions occur frequently in practice, e.g., where one aims to optimize the parameters of a system characterized by a Monte Carlo simulator that may only produce data samples from an intractable likelihood~\citep{cranmer2019frontier}. In such cases, genetic algorithms~\citep{opt_design, genetic_book}, Bayesian optimization~\citep{snoek2012practical, shahriari2015taking}, or numerical differentiation~\citep{moving_ass} are frequently employed. More recently stochastic gradient estimators~\citep{mohamed2019monte} such as the REINFORCE~\citep{Williams_reinforce} estimator have been employed to estimate gradients of non-differentiable functions~\citep{Stulp2013PolicyI,Chen2016LearningTL} and subsequently perform gradient-based optimization.

In order to utilize the strengths of gradient-based optimization while avoiding the high variance often observed with score function gradient estimators, our approach employs deep generative models as differentiable surrogate models to approximate non-differentiable simulators, as described in Figure~\ref{Fig:algo_graph}. Using these surrogates, we show that we can both approximate the stochastic behavior of the simulator and enable direct gradient-based optimization of an objective by parameterizing the surrogate model with the relevant parameters of the simulator. In high-dimensional parameter spaces, training such surrogates over the complete parameter space becomes computationally expensive. Our technique, which we name \emph{local generative surrogate optimization} (L-GSO), addresses this by using successive local neighborhoods of the parameter space to train surrogates at each step of parameter optimization. Our method works especially well when the parameters, which are seemingly high dimensional, live on a lower dimensional submanifold, as seen in practice in a variety of settings~\citep{hoos2014efficient}.

L-GSO relies primarily on two assumptions: (a) that the objective function is continuous and differentiable, and (b) that the parameters $\bpsi$ are continuous variables. The first assumption may be relaxed by incorporating the objective into the surrogate.

In Section~\ref{method} we describe the L-GSO algorithm. We cover related work in Section~\ref{sec:related_work}. In Section~\ref{sec:experiments} we evaluate L-GSO on a set of toy problems and compare it with frequently used methods including numerical differentiation, Bayesian optimization, and score function-based approaches, and present results of a realistic use case in the high-energy physics domain. Section~\ref{sec:conclusions} presents our conclusions.

\section{Method} \label{method}
\paragraph{Problem Statement}
We target an optimization formulation applicable in domains where a simulator characterized by parameters $\bpsi$ takes stochastic inputs $\bx \sim q(\bx)$ and produces outputs (observations) $\by \sim p(\by | \bx; \bpsi)$. For example in the case of designing the shape of an experimental device, $\bx$ may represent random inputs to an experimental apparatus, $\bpsi$ define the shape of the apparatus, and $p(\by | \bx; \bpsi)$ encodes the impact of the apparatus on the input to produce observations $\by$. A task-specific objective function $\mathcal{R}(\by)$ encodes the quality of observations and may be optimized over the parameters $\bpsi$ of the observed distribution. In cases when a simulator $F$ can only draw samples from the distributions $p(\by | \bx; \bpsi)$ the optimization problem can be approximated as
\begin{equation}
\begin{aligned}\label{eq:problem}
\bpsi^{*} = \argmin_{\bpsi} \E[\mathcal{R}(\by)] &= \argmin_{\bpsi} \int \mathcal{R}(\by) p(\by | \bx; \bpsi) q(\bx) d\bx d\by \\
& \approx \argmin_{\bpsi} ~ \frac{1}{N} \sum_{i=1}^{N} \mathcal{R}(F(\bx_i; \bpsi))\;
\end{aligned}
\end{equation}
where $ \by_i = F(\bx_i; \bpsi) \sim p(\by | \bx; \bpsi),~~ x_i \sim q(\bx)$ and a Monte Carlo approximation to the expected value of the objective function is computed using samples drawn from the simulator.  Note that $F$ represents a stochastic process, which may itself depend on latent random variables.

\subsection{Deep generative models as differentiable surrogates}

Given a non-differentiable simulator $F$, direct gradient-based optimization of \Eqref{eq:problem} is not possible. We propose to approximate $F$ with a learned differentiable model, denoted a surrogate, $\bar{\by} = S_{\theta} (\bz, \bx; \bpsi)$ that approximates $F(\bx; \bpsi)$, where $\bz \sim p(\bz)$ are latent variables accounting for the stochastic variation of the distribution $p(\by | \bx ; \bpsi)$, $\theta$ are surrogate model parameters, and $\bar{\by}$ are surrogate outputs. When the samples $\bar{\by}$ are differentiable with respect to $\bpsi$, direct optimization of \Eqref{eq:problem} can be done with the surrogate gradient estimator:
\begin{equation}\label{eq:sgrad}
\nabla_{\boldsymbol{\psi}} \E[\mathcal{R}(\by)] \approx
\frac{1}{N} \sum_{i=1}^N \nabla_{\boldsymbol{\psi}} \mathcal{R}(S_{\theta} (\bz_i, \bx_i; \boldsymbol{\psi}))\;.
\end{equation}

To obtain a differentiable surrogate capable of modeling a stochastic process, $S_\theta$ is defined as a deep generative model whose parameters $\theta$ are learned. Generative model learning can be done independently of the simulator optimization in~\Eqref{eq:problem} as it only requires samples from the simulator to learn the stochastic process. Once learned, the generative surrogate can produce differentiable samples that are used to approximate the integration for the expected value of the objective function.  Several types of generative models can be used, including generative adversarial networks (GANs)~\citep{gan_paper}, variational autoencoders~\citep{vae_well,vae_rezende}, or flow based models~\citep{rezende_nf}.  We present results using conditional variants of two recently proposed models, Cramer GAN~\citep{cramer_gan} and the FFJORD continuous flow model~\citep{ffjord_paper}, to show the independence of L-GSO from the choice of generative model.

\subsection{Local generative surrogates}

\begin{wrapfigure}{R}{0.5\textwidth}
\vspace{-2 \baselineskip}
\begin{minipage}{0.5\textwidth}
\begin{algorithm}[H]
\small
\caption{Local Generative Surrogate Optimization (L-GSO) procedure}\label{alg:euclid_local}
\begin{algorithmic}[1]
\REQUIRE number N of $\bpsi$, number M of $\bx$ for surrogate training, number K of $\bx$ for $\bpsi$ optimization step, trust region $U_{\epsilon}$, size of the neighborhood $\epsilon$, Euclidean distance $d$
\STATE \text{Choose initial parameter $\bpsi$}
\WHILE{$\bpsi$ has not converged}
\STATE \text {Sample $\bpsi_i$ in the region $U_{\epsilon}^{\bpsi}, \ i=1, \dots, N$} \label{sample_algo}
\STATE \begin{varwidth}[t]{\linewidth}
            \text{For each $\bpsi_i$, sample inputs $\{\bx^i_j\}_{j=1}^M \sim q(\bx)$}
\end{varwidth}
\STATE \begin{varwidth}[t]{\linewidth}
            \text{Sample $M \times N$ training examples from} \par \text{simulator}
             $\by_{ij} = F( \bx^i_j; \bpsi_i)$
        \end{varwidth}
\STATE \begin{varwidth}[t]{\linewidth}
    \text {Store $\by_{ij}, \bx^i_j, \bpsi_i$ in history $H$} \par
    $i=1, \dots, N; j=1, \dots, M$
    \end{varwidth}
\STATE \begin{varwidth}[t]{\linewidth}
            \text {Extract all $\by_{l}, \bx_l, \bpsi_l$ from history $H$,} \par
            \text{iff $d(\bpsi, \bpsi_l) < \epsilon$}
        \end{varwidth}
\STATE \begin{varwidth}[t]{\linewidth}
            \text{Train generative surrogate model} \par
            \text{$S_{\theta} (\bz_l, \bx_l; \bpsi_l)$, where $\bz_l \sim \mathcal{N}(0,1)$}
            \label{local_sur_train_line}
        \end{varwidth}
\STATE \text{Fix weights of the surrogate model} $\theta$

\STATE \begin{varwidth}[t]{\linewidth}
            \text{Sample} $\bar{\by}_k = S_{\theta} (\bz_k, \bx_k; \bpsi), \bz_k \sim \mathcal{N}(0,1)$,\par
            $\bx_k \sim q(\bx), \ k=1, \dots, K$
        \end{varwidth}

\STATE $\nabla_{\bpsi} \E[\mathcal{R}(\bar{\by})] \leftarrow \frac{1}{K} \sum\limits_{k=1}^{K} \frac{\partial \mathcal{R}}{\partial{\bar{\by}_k}} \frac{\partial S_{\theta} (\bz_k, \bx_k; \bpsi)}{\partial \bpsi}$

\STATE $\boldsymbol{\psi} \leftarrow \text{SGD} (\psi, \nabla_{\boldsymbol{\psi}} \E[\mathcal{R}(\bar{\by})])$
\ENDWHILE
\end{algorithmic}
\label{opt_algo_local}
\end{algorithm}
\end{minipage}
\vspace{- \baselineskip}
\end{wrapfigure}

The L-GSO optimization algorithm is summarized in Algorithm~\ref{opt_algo_local}. Using a sample of values for $\bpsi$ and input samples of $\bx$, a set of training samples for the surrogate are created from the simulator $F$. The surrogate training step 8 refers to the standard training procedures for the chosen generative model (details on model architectures and hyperparameters are given in Appendix~\ref{sur_get_appen}). The learned surrogate is used to estimate the gradient of the objective function with backpropagation through the computed expectation of ~\Eqref{eq:sgrad} with respect to $\bpsi$.  Subsequently $\bpsi$ is updated with a gradient descent procedure, denoted \textit{SGD} (stochastic gradient descent) in the algorithm. Due to the use of SGD, an inherently noisy optimization algorithm, the surrogate does not need to be trained until convergence but only sufficiently well to provide gradients correlated with the true gradient that produce useful SGD updates. The level of correlation will control the speed of convergence of the method.

For high-dimensional $\bpsi$, a large number of parameter values $\bpsi$ must be sampled to accurately train a single surrogate model. Otherwise the surrogate would not provide sufficiently well estimated gradients over the full parameter space that may be explored by the optimization. Thus optimization using a single upfront training of the surrogate model over all $\bpsi$ becomes unfeasible.  As such, we utilize a trust-region like approach~\citep{trustRegion} to train a surrogate model locally in the proximity of the current parameter value $\bpsi$. We sample new $\bpsi^{\prime}$ around the current point $\bpsi$ from the set $U_{\epsilon}^{\bpsi} = \{\bpsi^{\prime}; | \bpsi^{\prime} - \bpsi |  \leq \epsilon \}$. Using this local model, a gradient at the current point $\bpsi$ can be obtained and a step of SGD performed. After each SGD update of $\bpsi$, a new local surrogate is trained. As a result, we do not expect domain shift to impact L-GSO as it is retrained at each new parameter point.

In local optimization there are several hyperparameters that require tuning either prior to or dynamically during optimization. One must choose the sampling algorithm for $\bpsi$ values in the region $U_{\epsilon}^{\psi}$ in step~3 of Algorithm~\ref{opt_algo_local}. In high-dimensional space, uniform sampling is inefficient, thus we have adopted the Latin Hypercubes algorithm~\citep {lhc_sampling}.\footnote{A detailed study of alternative sampling techniques is left for future work.} One must also choose a proximity hyperparameter $\epsilon$, that controls the size of the region of $\boldsymbol{\psi}_i$ in which a set of $\bpsi$ values is chosen to train a local surrogate.\footnote{For a deterministic simulator this parameter could be chosen proportionally to learning rate. If the objective function is stochastic, one might want to choose $\epsilon$ big enough so that $\E |\mathcal{R}(\mathbf{y}_{\psi - \epsilon}) - \mathcal{R}(\mathbf{y}_{\psi + \epsilon})) | > \mathrm{Var}(\mathcal{R}(\mathbf{y}_{\psi}))$.} This hyperparameter is similar to the step size used in numerical differentiation, affecting the speed of convergence as well as the overall behavior of the optimization algorithm; if $\epsilon$ is too large or too small the algorithm may diverge. In this paper we report experimental results with this hyperparameter tuned based on a grid search. The value of $\epsilon=0.2$ was found to be robust and work well in all experiments except the "three hump problem" which required a slightly larger value of $\epsilon = 0.5$. More details can be found in Appendix~\ref{appendix_opt_details}.

The number of $\bpsi$ values sampled in the neighborhood is another key hyperparameter. We expect the optimal value to be highly correlated with the dimensionality and complexity of the problem. In our experiments, we examine the quality of gradient estimates as a function of the number of points used for training the local surrogate. We observe that it is sufficient to sample $O(D)$ samples in the neighborhood of $\boldsymbol{\psi}$, where $D$ is the full parameter space dimensionality of $\boldsymbol{\psi}$. In this case, our approach is observed to be similar to numerical optimization which expects $O(D)$ samples for performing a gradient step~\citep{moving_ass}. However, in cases where the components of $\bpsi$ relevant for the optimization lie on a manifold of dimensionality lower than $D$, i.e., intrinsic dimensionality $d$ is smaller than $D$, we observe that L-GSO requires less than $D$ samples for producing a reasonable gradient step, thus leading to the faster convergence of L-GSO than other methods. 

Previously sampled data points can also be stored in history and later reused in our local optimization procedure (Algorithm~\ref{opt_algo_local}). This provides additional training points for the surrogate as the optimization progresses. This results in a better surrogate model and, consequently, better gradient estimation. The ability of L-GSO to reuse previous samples is a crucial point to reduce the overall number of calls to the simulator. This procedure was observed to aid both FFJORD and GAN models to attain the minimum faster and to prevent the optimization from diverging once the minimum has been attained.

The benefit of our approach, in comparison with numerical gradient estimation, is that a deep generative surrogate can learn more complex approximations of the objective function than a linear approximation, which can be beneficial to obtain gradients for surfaces with high curvature. In addition, our method allows a reduction of the number of function calls by reusing previously sampled points. Utilizing generative neural networks as surrogates also provides potential benefits such as Hessian estimation, that may be used for second-order optimization algorithms and/or uncertainty estimation, and possible automatic determination of a low-dimensional parameter manifold.

\section{Related work}
\label{sec:related_work}
Our work intersects with both simulator optimization and likelihood-free inference. In terms of simulator optimization, our approach can be compared to Bayesian optimization (BO) with Gaussian process based surrogates~\citep{snoek2012practical, shahriari2015taking} and numerical derivatives~\citep{moving_ass}. In comparison with BO, our optimization approach makes use of gradients of the objective surface approximated using a surrogate, which can result in a faster and more robust convergence in multidimensional spaces with high curvature (see the Rosenbrock example in Figure~\ref{mean_c}). Importantly, our approach does not require covariance matrix inversion which costs $O(n^3)$ where $n$ is a number of observations, thus making it impractical to compute in high dimensional spaces. To make BO scalable, authors often make structural assumptions on the function that may not hold generally. For example, references~\citep{wang2013bayesian, djolonga2013high,10.5555/3020751.3020776,Zhang2019HighDB}, assume a low-dimensional linear subspace that can be decomposed in subsets of dimensions. In addition, BO may require the construction of new kernels~\citep{gp_kernel_selection}, such as \citep{Oh2018BOCKB} which proposes a cylindrical kernel that penalizes close to boundary points in the search space. Our approach does not make structural assumptions of the parameter manifold or assumptions on the locality of the optimum. While our approach does not require the design of a task-specific kernel, it does require the selection of a surrogate model structure. The method of reference~\citep{NIPS2019_8788} maintains multiple local models simultaneously and samples new data points via an implicit multi-armed bandit approach.

In likelihood-free inference, one aim is to estimate the parameters of a generative process with neither a defined nor tractable likelihood. A major theme of this work is posterior or likelihood density estimation~\citep{Papamakarios2018SequentialNL,Lueckmann2018LikelihoodfreeIW,Greenberg2019AutomaticPT}. Our work is similar in its use of sequentially trained generative models for density estimation, but we focus on optimizing any user-defined function, and not specifically a likelihood or posterior, and our sequential training is used to enable gradient estimation rather than updating a posterior.  Other recent examples of work that address this problem include~\citep{Louppe2017AdversarialVO,grathwohl2018backpropagation, ruiz2018learning}. While the first reference discusses non-differentiable simulators, it targets tuning simulator parameters to match a data sample and not the optimization of a user-defined objective function.

Non-differentiable function optimization using score function gradient estimators is explored in~\citep{grathwohl2018backpropagation, ruiz2018learning}. These approaches are applicable to our setting and we provide comparison in our experiment. Instead of employing the simulator within the computation of score function gradient estimate, our approach builds a surrogate simulator approximation to estimate gradients.

\begin{wrapfigure}{l}{0.5\textwidth}
\includegraphics[width=1\linewidth]{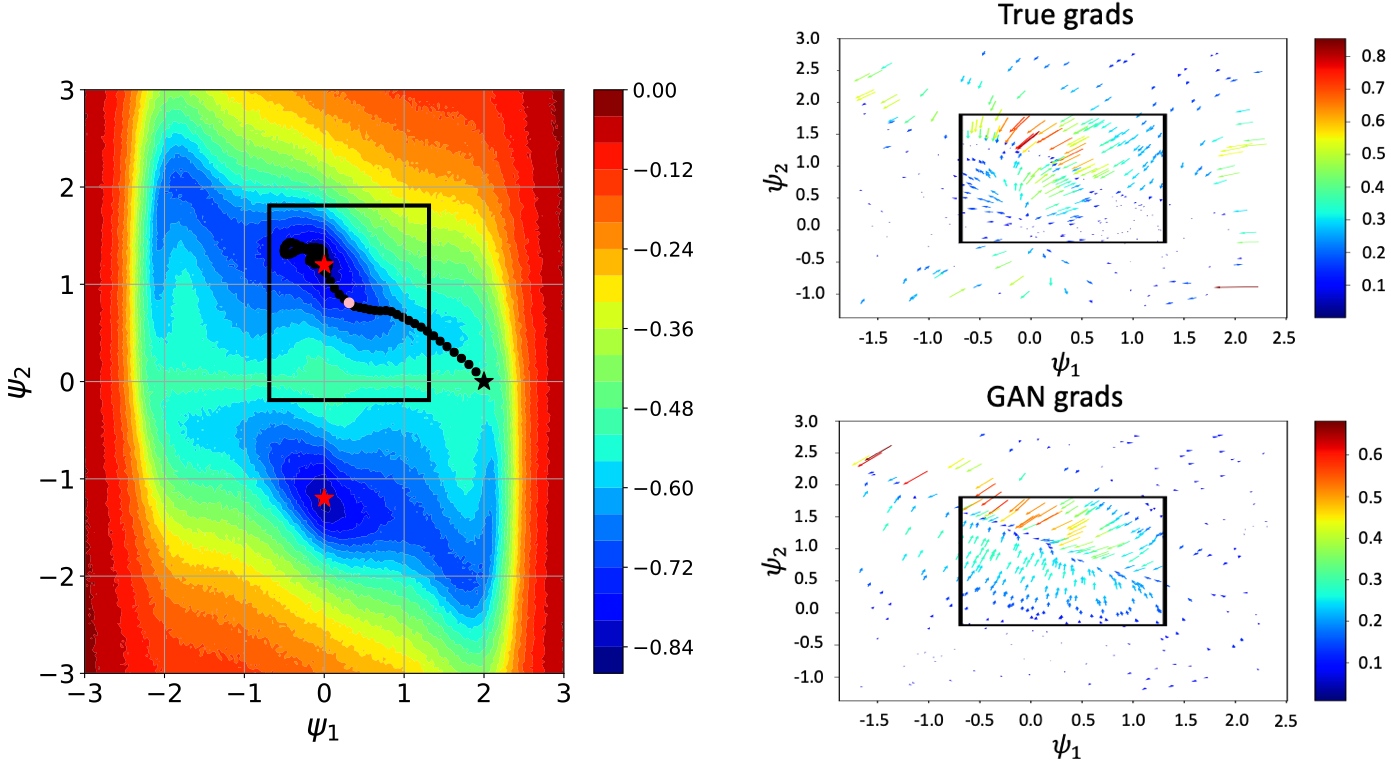}
\caption{(Left) objective function surface of the "hump model" overlaid by the optimization path. Red stars are the objective optimal values. (Right) True gradients and GAN gradients, calculated at the yellow point. Black rectangle correspond to the current $\epsilon$ neighborhood around yellow point. Full path animation is available at \url{https://doi.org/10.6084/m9.figshare.9944597.v3}.}
\label{path_plots}
\end{wrapfigure}%

Deep generative model for optimization problems have been explored in~\citep{Gupta2019FeedbackGF,Brookes2019ConditioningBA}. These approaches focus on posterior inference of the parameters given observations, whereas we focus on direct gradient based objective optimization. These algorithms also assume that it is not costly to sample from the simulator to perform posterior estimation, while sampling from the simulator is considered to be the most computationally expensive step to be minimized in our approach. In addition,~\citep{GmezBombarelli2018AutomaticCD} use generative models to learn a latent space where optimization is performed, rather than direct optimization over the parameter space of a surrogate forward model as in our approach.

The success of L-GSO in high-dimensional spaces with low-dimensional submanifolds is consistent with findings on the approximation quality of deep networks and how they adapt to the intrinsic dimension of data~\citep{Suzuki2019DeepLI,Nakada2019AdaptiveAA,SchmidtHieber2019DeepRN}. Although restricted to specific function classes, these results provide bounds on approximation quality that reduce with small intrinsic dimension. They are suggestive of the benefits of deep generative models over submanifolds that we empirically observe.

\section{Experiments}
\label{sec:experiments}

We evaluate L-GSO on five toy experiments in terms of the attained optima and the speed of convergence, and present results in a physics experiment optimization. As simulation is assumed to be the most time consuming operation during optimization, the speed of convergence is measured by the number of simulator calls. The toy experiments, defined below, were chosen to explore low- and high-dimensional optimization problems, and those with parameters on submanifolds. Non-stochastic versions of the experiments are established benchmark functions in the optimization literature~\citep{Jamil2013ALS}.

\vspace{-0.5em}
\paragraph{Probabilistic Three Hump Problem}\label{three_problem} We aim to find the 2-dimensional $\bpsi$ that optimizes:
\begin{equation}
\begin{gathered}
\bpsi^{*} = \argmin_{\bpsi} \E[\mathcal{R}(y)] = \E[\sigma(y - 10) - \sigma(y)],\; \textrm{s.t.} \\
y \sim \mathcal{N}\left(y; \mu_i, 1 \right), i \in \{1,2\}, ~~~ 
\mu_i \sim \mathcal{N}(x_{i} h(\boldsymbol{\psi}), 1), ~~~ x_1 \sim \mathrm{U}[-2, 0], ~~~ x_2 \sim \mathrm{U}[2, 5]\\
\prob{i=1} = \frac{\psi_1}{||\boldsymbol{\psi}||_2} = 1 - \prob{i=2}, ~~~ h(\boldsymbol{\psi}) = 2 \psi_1^2 - 1.05 \psi_1^4 + \psi_1^6 / 6 + \psi_1 \psi_2 + \psi_2^2
\end{gathered}
\end{equation}

\paragraph{Rosenbrock Problem} In the N-dimensional Rosenbrock problem we aim to find $\bpsi$ that optimizes:
\begin{gather}
\begin{gathered}
\bpsi^{*} = \argmin_{\bpsi} \E[\mathcal{R}(y)] = \argmin_{\bpsi} \E[y],\;\textrm{s.t.} \\
y \sim \mathcal{N}\left(y; \sum\limits_{i=1}^{n - 1} \left[(\psi_i - \psi_{i+1})^2 + (1 - \psi_i)^2\right] + x, 1\right),~~~ 
x \sim \mathcal{N}(x; \mu, 1),~~~\mu \sim \mathrm{U}[-10, 10]
\end{gathered}
\end{gather}

\paragraph{Submanifold Rosenbrock Problem}
\label{degen_problem}
To address problems where the parameters lie on a low-dimension submanifold, we define the submanifold Rosenbrock problem, with a mixing matrix $A$ to project the parameters onto a submanifold.\footnote{\label{qr_mixing_matrix}The orthogonal mixing matrix $A$ is generated via a QR-decomposition of a random matrix sampled from the normal distribution.} In our experiments $\bpsi \in \mathbb{R}^{100}$ and $A \in \mathbb{R}^{10 \times 100}$ has full row rank. Prior knowledge of $A$ or the submanifold dimension is not used in the surrogate training. The optimization problem is thus defined as:
\begin{gather}
%\begin{gathered}
\bpsi^{*} = \argmin_{\bpsi} \E[\mathcal{R}(y)] = \argmin_{\bpsi} \E[y], ~\textrm{s.t.}~
y \sim \mathcal{N}\left(y; \sum\limits_{i=1}^{n - 1} \left[(\psi_i' - \psi_{i+1}')^2 + (1 - \psi_i')^2\right] + x, 1\right) \nonumber \\
\boldsymbol{\psi}' = A \cdot (\boldsymbol{\psi}[\mathrm{mask}]),~~
x \sim \mathcal{N}(x; \mu, 1),~~\mu \sim \mathrm{U}[-10, 10]
%\end{gathered}
 \end{gather}

\paragraph{Nonlinear Submanifold Hump Problem}
\label{degen_problem_hump}
This experiment explores non-linear submanifold embeddings of the parameters $\psi$. The embedding is through $\hat{\psi} = B\tanh(A \psi)$, where $\psi \in \mathbb{R}^{40}$, $A \in \mathbb{R}^{16 \times 40}$, $B \in \mathbb{R}^{2 \times 16}$, with $A$ and $B$ generated as in the submanifold Rosenbrock problem.\footnoteref{qr_mixing_matrix} The intrinsic dimension of the parameter space is equal to two. The embedded parameters $\hat{\psi}$ are used in place of $\psi$ in the three hump problem definition. We use this example to also explore the number of parameter points needed per surrogate training for effective optimization on a submanifold.

\paragraph{Neural Network Weights Optimization Problem}
\label{boston_sec}
In this problem, we optimize neural network weights for regressing the Boston house prices dataset~\citep{bostonD}. As discussed by~\citep{Li2018MeasuringTI}, neural networks are often overparameterized, thus having a smaller intrinsic dimensions than the full parameter space dimension. In this experiment we explore the optimization capability of L-GSO over the number of parameter space points needed per surrogate training, and, indirectly, the intrinsic dimensionality of the problem. The problem is defined as:
\begin{gather}
\begin{gathered}
\bpsi^{*} = \argmin_{\bpsi} \E[\mathcal{R}(y)] = \argmin_{\bpsi} \sqrt{\frac{1}{N} \sum_{i=1}^N (y - y_{\mathrm{true}})^2}, \textrm{s.t.} ~~
y = \mathrm{NN}(\bpsi, \bx), \bx \sim \{ \bx_i \}_{i=1}^{506}
\end{gathered}
\end{gather}

\paragraph{Baselines}
We compare L-GSO to: Bayesian optimization using Gaussian processes with cylindrical kernels \citep{Oh2018BOCKB}, which we denote ``\textit{BOCK}'', numerical differentiation with gradient descent (referred to as numerical optimization), and guided evolutionary strategies~\citep{Maheswaranathan2018GuidedES}.

\begin{figure*}[t]
\begin{center}
\begin{subfigure}[h]{0.31\linewidth}
\includegraphics[width=\linewidth]{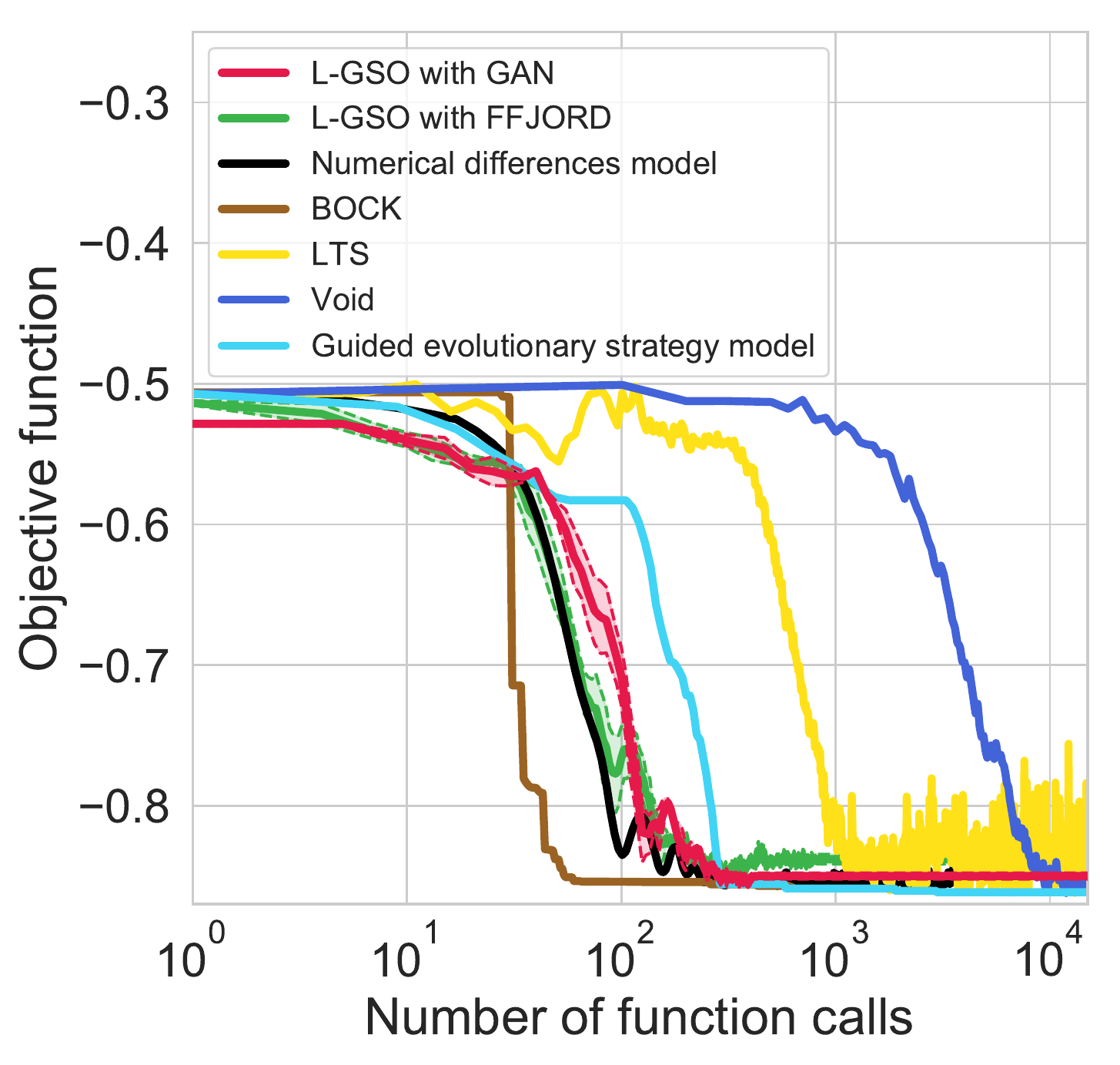}
\caption{}
\label{mean_a}
\end{subfigure}\quad
\begin{subfigure}[h]{0.31\linewidth}
\includegraphics[width=\linewidth]{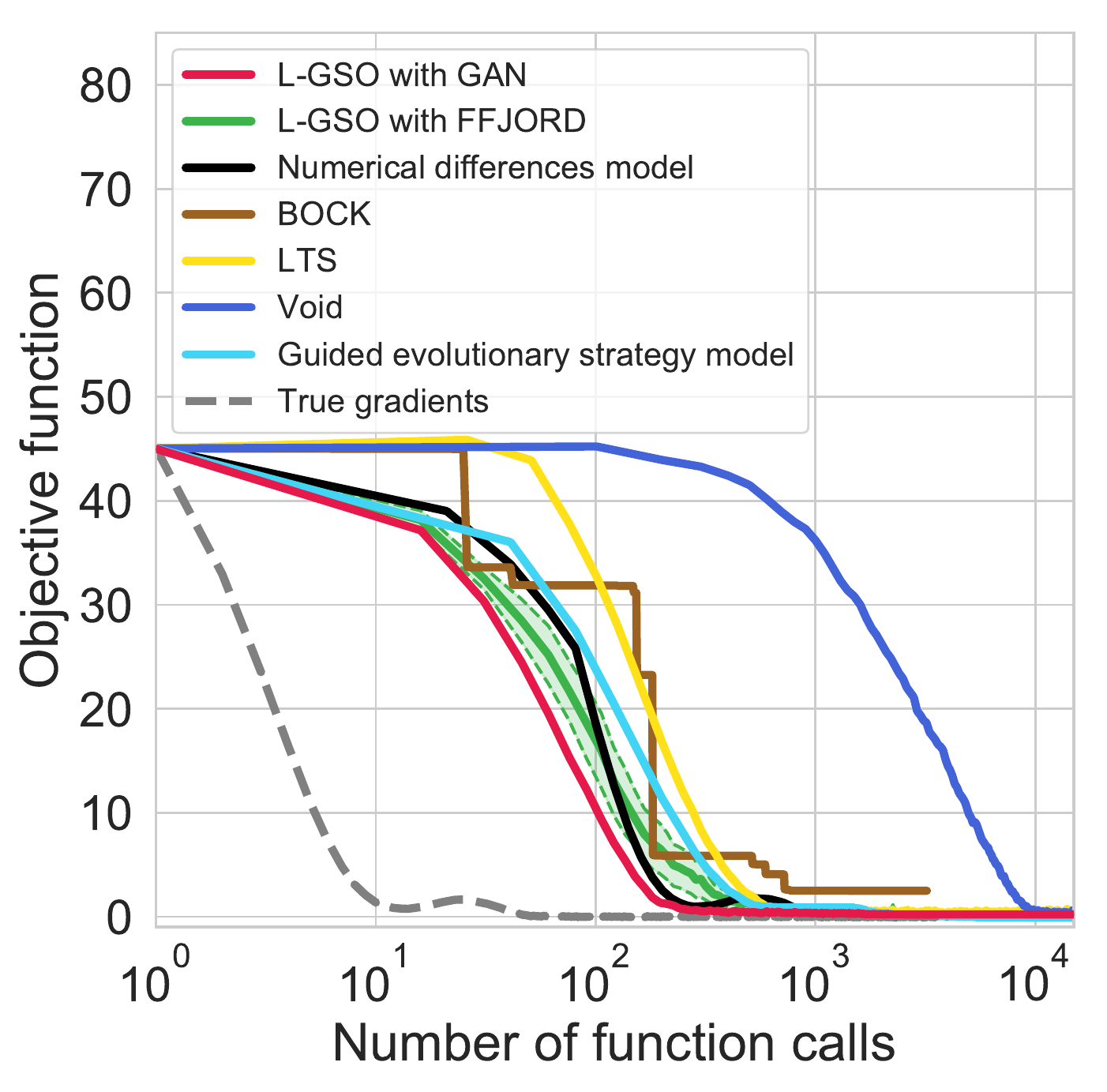}
\caption{}
\label{mean_b}
\end{subfigure}\quad
%\newline
\begin{subfigure}[h]{0.31\linewidth}
\includegraphics[width=\linewidth]{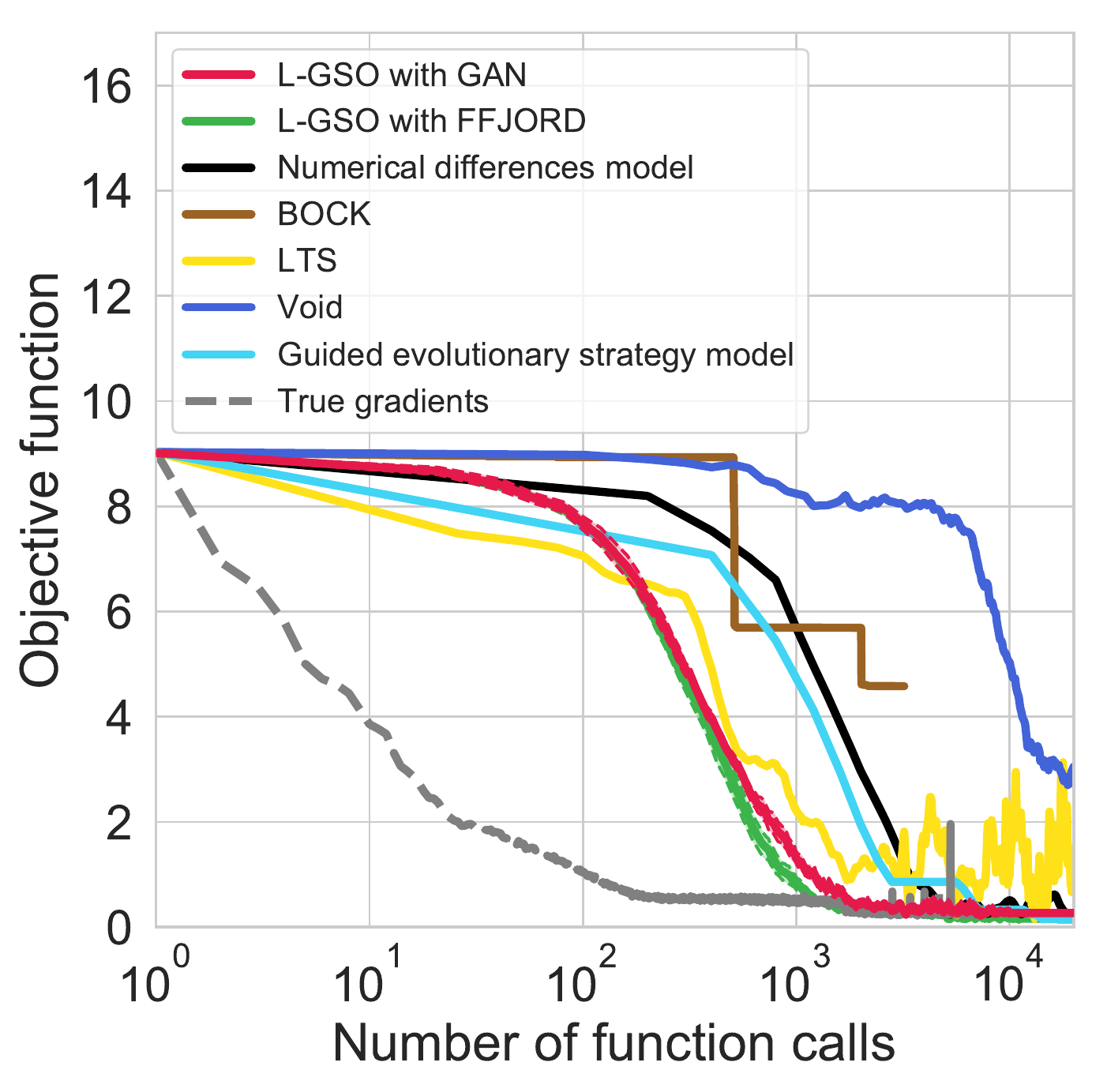}
\caption{}
\label{mean_c}
\end{subfigure}
\end{center}
\caption{The objective function value on the toy problems for baselines and our method. (a) Three hump problem,
(b) Rosenbrock problem~\citep{Rosenbrock} in 10 dimensions, initial point is $\vec{2} \in \mathbb{R}^{10}$.  (c) Submanifold Rosenbrock Problem in 100 dimensions, initial point is $\vec{2} \in \mathbb{R}^{100}$. True gradients are shown in gray dashed curves when available. Shaded region corresponds to $1\sigma$ confidence intervals.}
\label{fig:mean}
\end{figure*}%

We also compare with score function-based optimization approaches.  In "Learning to Simulate"~\citep{ruiz2018learning}, in order to apply score function gradient estimation of a non-differentiable simulator, the authors introduce a policy over the simulator parameters, $p(\psi | \eta)$, and optimize the policy parameters $\eta$. We denote this method ``LTS". Reference~\citep{grathwohl2018backpropagation} introduces the LAX gradient which uses an optimized control variate to reduce the variance of the score function gradient estimator. We cannot apply this method directly, as the authors optimize objectives of form $L(\psi) = \E_{p(y|\psi)}[f(y)]$ and use knowledge of the gradient $\nabla_{\psi} \log p(y|\psi)$, and in our setting with a non-differentiable simulator this gradient is not available. Following~\citep{ruiz2018learning}, we treat the objective as a function of the parameters, $f(\psi) = \E_{p(\by| \bx; \bpsi)}[\mathcal{R}(\by)]$, introduce a Gaussian policy $p(\psi | \mu, \sigma)=N(\psi | \mu, \sigma)$, and optimize the objective $L(\mu, \sigma) = \E_{p(\psi | \mu, \sigma)}[f(\psi)]$ using LAX.

\begin{wrapfigure}{l}{0.6\textwidth}
\vspace{- \baselineskip}
\begin{subfigure}{0.49\linewidth}
\includegraphics[width=\linewidth]{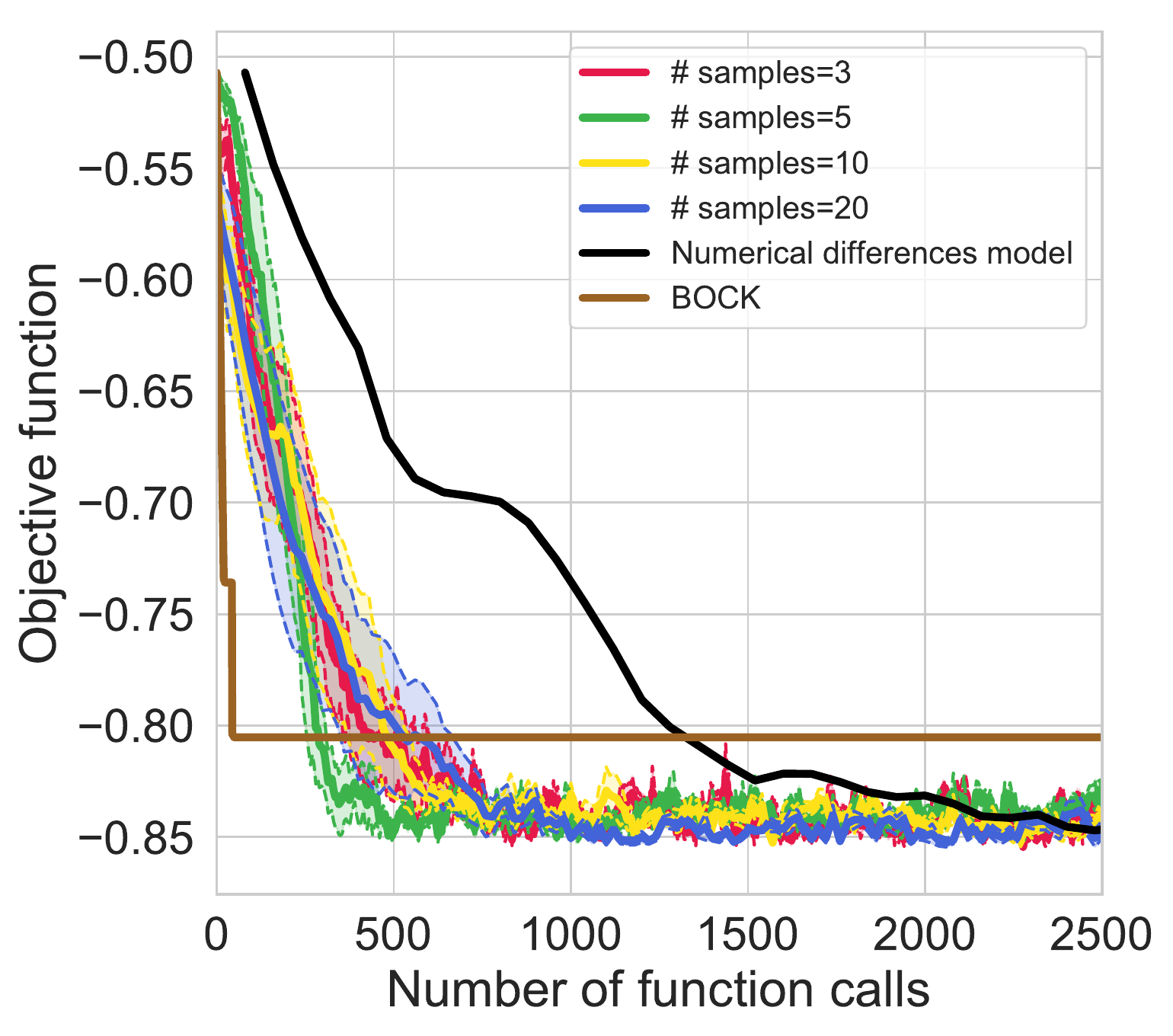}
\caption{}
\label{manifold_a}
\end{subfigure}
\begin{subfigure}{0.49\linewidth}
\includegraphics[width=0.9\linewidth]{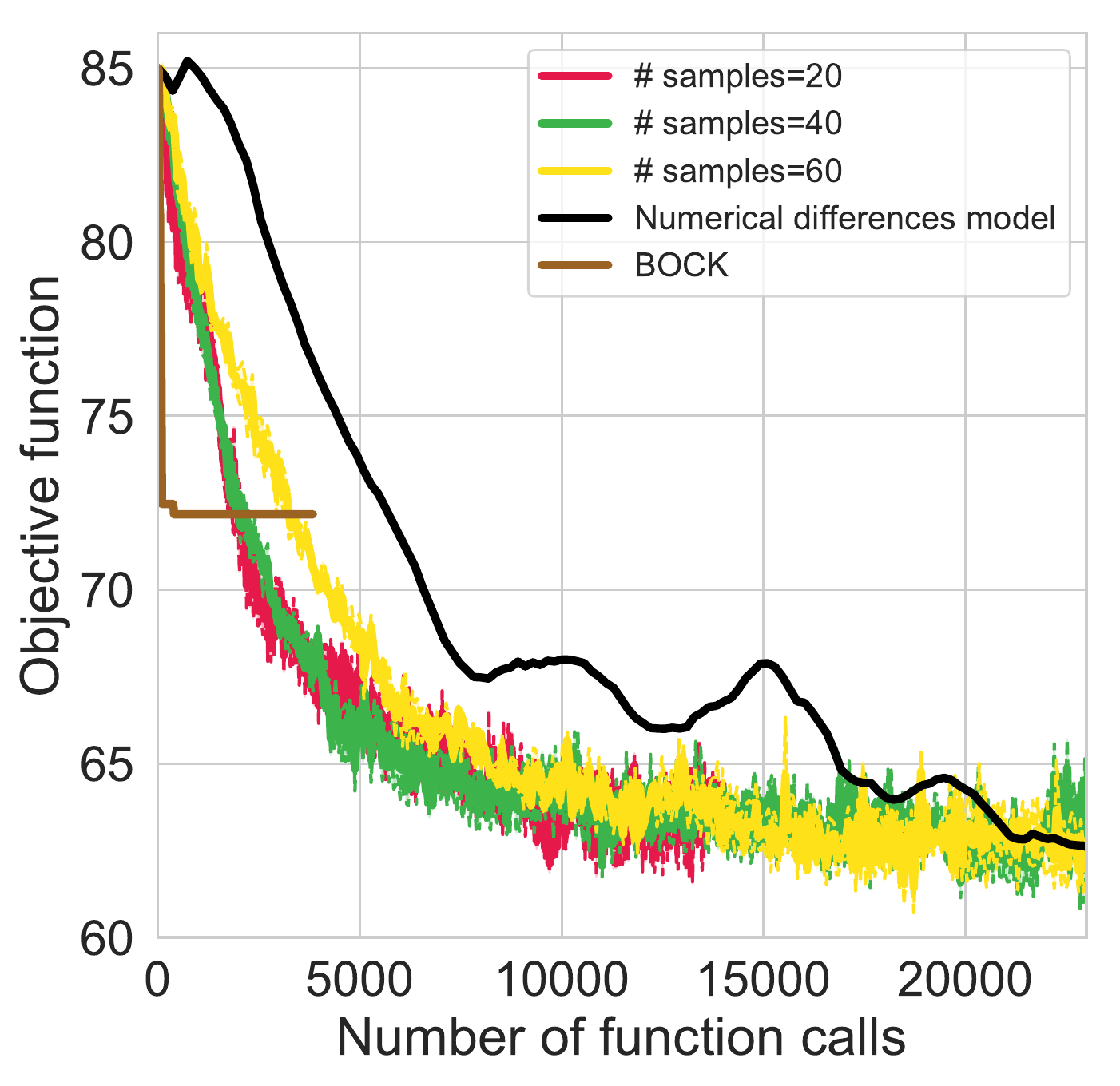}
\caption{}
\label{manifold_b}
\end{subfigure}%
\caption{The objective function value as a function of the accumulated number of simulator calls for (a) Nonlinear Submanifold Three Hump problem, $\bpsi \in \mathbb{R}^{40}$,
(b) Neural Network Weights Optimization problem , $\bpsi \in \mathbb{R}^{91}$.}
\label{fig:manifold}
\end{wrapfigure}

Our primary metrics for comparison are the objective function value obtained from the optimization, and the number of simulator function calls needed to find a minimum. The latter metric assumes that the simulator calls are driving the computation time of the optimization. We tuned the hyper-parameters of all baselines for their best performance. For all experiments we run L-GSO five times with different random seeds and the variance is shown as bands in Figures \ref{fig:mean} and \ref{fig:manifold}. When presented, true gradients of the objective function are calculated with PyTorch \citep{NEURIPS2019_9015} and Pyro~\citep{Pyro}.

\vspace{-1em}
\paragraph{Results}
\label{results}
\begin{wrapfigure}{R}{0.4\textwidth}
\vspace{- \baselineskip}
\centering
\includegraphics[width=1\linewidth]{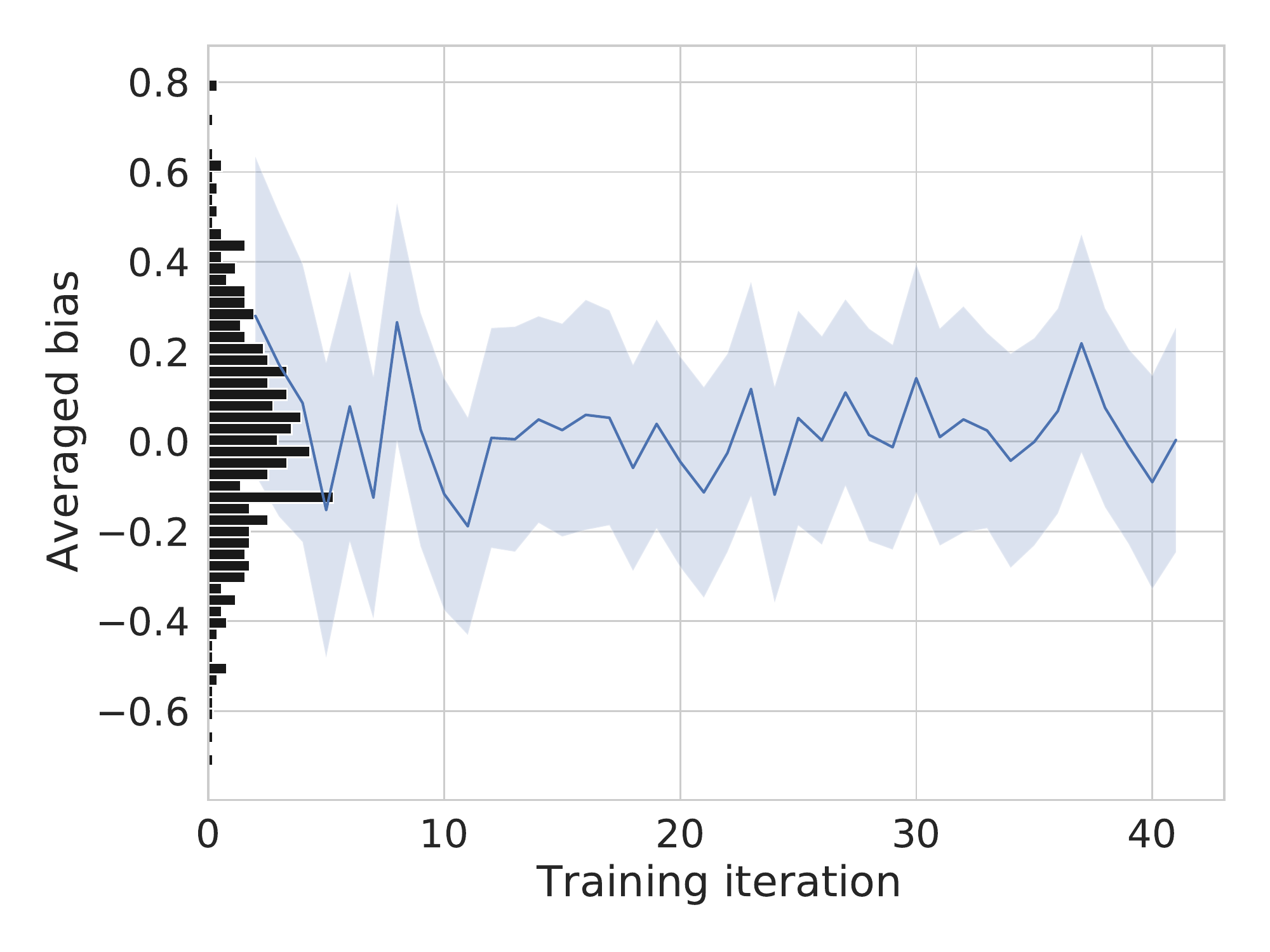}
\caption{The bias (solid line) and one standard deviation (shaded region) of the GAN based L-GSO gradient averaged over all $\bpsi$ dimensions in the 10D Rosenbrock problem versus training step. Gray histogram shows the empirical bias distribution over all training iterations.}
\label{bias_path}
\end{wrapfigure}%
To illustrate L-GSO, we show the optimization path in $\bpsi$ space for three hump problem in Figure \ref{path_plots}. We also show gradient vector fields of the true model and of the GAN surrogate estimation at random $\bpsi$ points during optimization, showing the successful non-linear approximation of the gradients by the surrogate. Visually, the true and the surrogate gradients closely match each other inside the surrogate training neighborhood (black rectangle). 

The objective value as a function of the number of simulator calls in three experiments is seen in Figure~\ref{fig:mean}.  L-GSO outperforms score function based algorithms in speed of convergence by approximately an order of magnitude. L-GSO also attains the same optima as other methods and the speed of convergence is comparable to numerical optimization. In Figure~\ref{mean_a}, BO converges faster than all other methods, which is not surprising in such a low-dimensional problem. Conversely, in Figure~\ref{mean_b} BO struggles to find the optimum due to the high curvature of the objective function, whereas the convergence speed of L-GSO is on par with numerical optimization. In general, L-GSO has several advantages over BO: (a) it is able to perform optimization without specification of the search space~\citep{NIPS2019_9350,pmlr-v51-shahriari16}, 
%i.e., $\boldsymbol{\psi} \in \mathbb{R}^p$, 
(b) the algorithm is embarrassingly parallelizable, though it should be noted that BO parallelization is an active area of research~\citep{Wang2016ParallelBG}.

The bias and variance of the GAN based L-GSO gradient estimate averaged over all parameters for the 10D Rosenbrock problem for each training step can be seen in Figure~\ref{bias_path}. The bias is calculated as the average difference between the true gradient and trained surrogates gradients, where each surrogate is trained with independent initialization and parameter sets $\{\psi^{\prime}\}$ in the $\epsilon$-neighborhood of the current value $\bpsi$, for each training iteration. The variance is similarly computed over the set of trained surrogates gradients. The bias remains close to, and well within one standard deviation of, zero across the entire training. Bias and variance for each parameter, and additional details, can be found in the Appendix~\ref{bias_appendix}.

The benefits of L-GSO can further be seen in problems with parameter submanifolds, i.e., the Submanifold Rosenbrock, Nonlinear Submanifold Hump Problem and Neural Network Weights Optimization problems where the relevant $\psi$ parameters live on a latent low-dimensional manifold. No prior knowledge of the submanifold is used in the training and all dimensions of $\bpsi$ are treated equally for all algorithms. The objective value versus the number of simulator calls can be seen in Figures~\ref{mean_c} and~\ref{fig:manifold} where we observe that L-GSO outperforms all baseline methods.  We also observe that BO was frequently not able to converge on such problems. Additionally, in all experiments, numerical optimization and/or BOCK outperform the score function-based approaches under consideration.

Figure~\ref{fig:manifold} compares L-GSO with differing numbers of parameter space points used per surrogate training. In the submanifold problems, L-GSO converges fastest with far fewer parameter points than the full dimension of the parameter space.  This indicates that the surrogate is learning about the latent manifold of the data, rather than needing to fully sample the volume around a current parameter point. The conditional generative network appears to learn the correlations between different dimensions of $\bpsi$ and is also able to interpolate between different sampled points. This allows the algorithm to obtain useful gradients in $\bpsi$, while using far fewer samples than numerical differentiation.

\subsection{Physics experiment example} \label{physcs_exp}

\begin{wrapfigure}{R}{0.4\linewidth}
\vspace{-\baselineskip}
\centering
\includegraphics[width=1\linewidth]{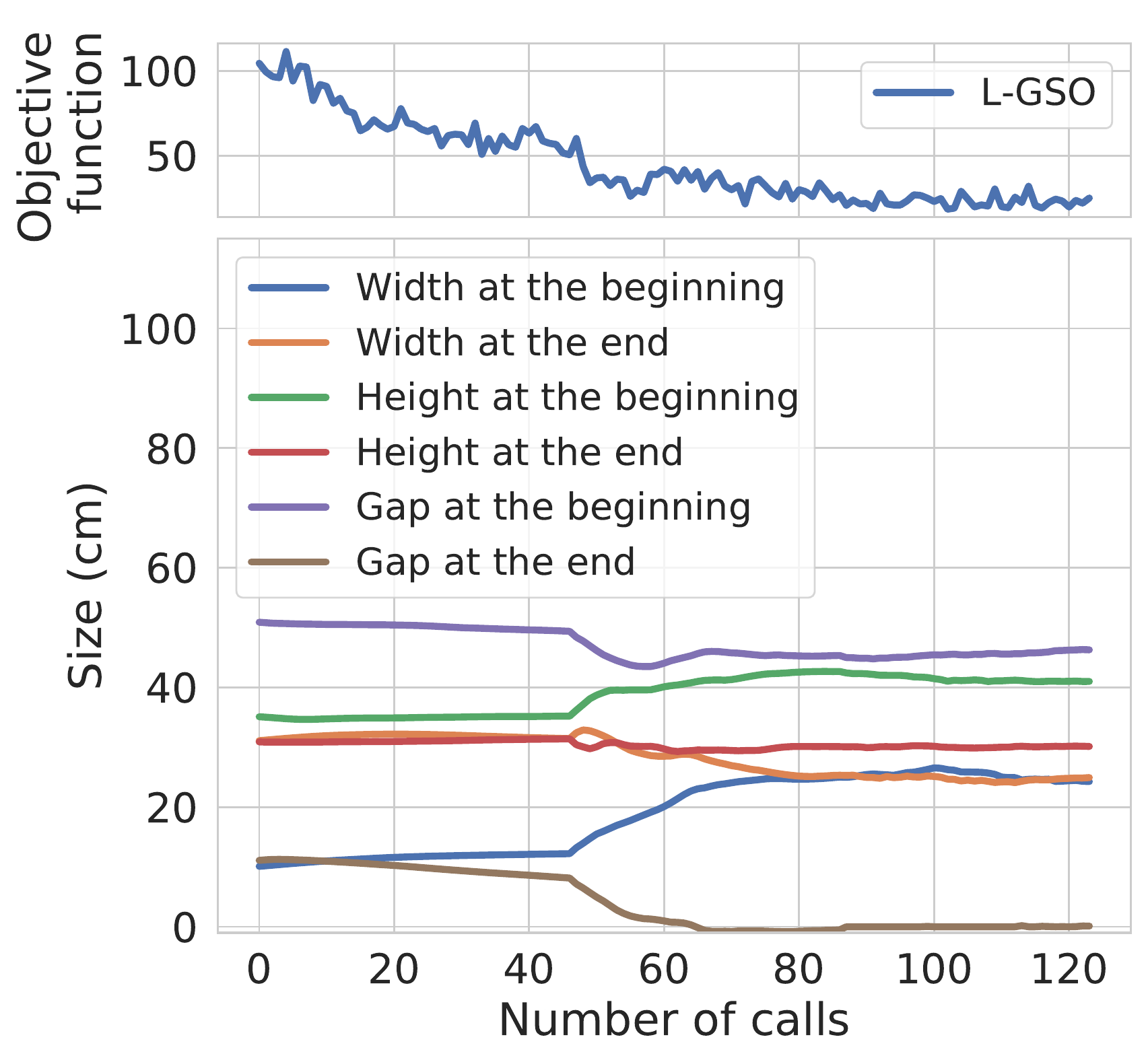}
\caption{Magnet objective function (top) and six $\bpsi$ parameters (bottom) during optimization.}
\label{ship_optimization}
\end{wrapfigure}%

We apply L-GSO to optimize the parameters of an apparatus in a physics experiment setting that uses the physics simulation software GEANT4~\citep{GEANT4} and FairRoot~\citep{FairRoot}. 
In this example, muon particles pass through a multi-stage steel magnet and their coordinates are recorded when muons leave the magnet volume if they pass through the sensitive area of a detection apparatus. As muons are unwanted in the experiment, the objective is to minimize number of recorded muons by varying the geometry of the magnet.

\paragraph{Problem Definition} Inputs $\bx$ correspond to the properties of incoming muons, namely the momentum ($\mathrm{P}$), the azimuthal ($\phi$) and polar ($\theta$) angles with respect to the incoming $z$-axis, the charge $Q$, and $x$-$y$-$z$ coordinate $C$. The observed output $\by$ is the muon coordinates on the plane transverse to the $z$-axis as measured by the detector. The parameter $\bpsi \in \mathbb{R}^{42}$ represents the magnet geometry. The objective function to penalize muons hitting the detector, where $\mathbbm{1}$ is the indicator function, and $\alpha_1 = 5.6$ m and $\alpha_2 = 3$ m define the detector sensitive region, is

\begin{gather}\nonumber
\mathcal{R}(\mathbf{y}; \boldsymbol{\alpha}) = \sum\limits_{i=1}^N \mathbbm{1}_{Q_i = 1}\sqrt{(\alpha_1 - (\by_i + \alpha_2)) / \alpha_1} + \mathbbm{1}_{Q_i = -1}\sqrt{(\alpha_1 + (\by_i - \alpha_2)) / \alpha_1}
\end{gather}

\textbf{Results} of the optimization using L-GSO with a Cramer GAN ~\citep{cramer_gan} surrogate are presented in Figure~\ref{ship_optimization}.  A previous optimization of this magnet system was performed using BO with Gaussian processes with RBF kernels~\citep{shield_opt_ship}. The L-GSO optima has an objective function value approximately 25\% smaller than the BO solution, while using approximately the same budget of $O(5,000)$ simulator calls. The L-GSO solution is shorter and has lower mass than the BO solution, which can both improve efficacy of the experiment and significantly reduce cost. More details can be found in the Appendix~\ref{Physics_problem_formulation}.

\section{Conclusions}
\label{sec:conclusions}

We present a novel approach for the optimization of stochastic non-differentiable simulators. Our proposed algorithm is based on deep generative surrogates successively trained in local neighborhoods of parameter space during parameter optimization. We compare against baselines including methods based on score function gradient estimators~\citep{ruiz2018learning,grathwohl2018backpropagation}, numerical differentiation, and Bayesian optimization with Gaussian processes \citep{Oh2018BOCKB}. Our method, L-GSO, is generally comparable to baselines in terms of speed of convergence, but is observed to excel in performance where simulator parameters lie on a latent low-dimensional submanifold of the whole parameter space. L-GSO is parallelizable, and has a range of advantages including low variance of gradient estimates, scalability to high dimensions, and applicability for optimization on high curvature objective function surfaces. We performed experiments on a range of toy problems and a real high-energy physics simulator. Our results improved on previous optimizations obtained with Bayesian optimization, thus showing the successful optimization of a complex stochastic system with a user-defined objective function.

\section*{Acknowledgments}

We would like to thank Gilles Louppe and Auralee Edelen for their feedback, and Fedor Ratnikov for the fruitful discussions.  MK is supported by the US Department of Energy (DOE) under grant DE-AC02-76SF00515 and by the SLAC Panofsky Fellowship. AU and VB are supported by the Russian Science Foundation under grant agreement n$^{\circ}$ 19-71-30020 for their work on Bayesian Optimization and FFJORD methods. AGB is supported by EPSRC/MURI grant EP/N019474/1 and by Lawrence Berkeley National Lab.

\bibliographystyle{plain}
\bibliography{bibliography}

\clearpage
\newpage

% \twocolumn[
% \center\Large
\begin{center}
\Large{\textbf{Black-Box Optimization with Local Generative Surrogates\\\vspace{1mm}
Supplementary Material}\\\vspace{5mm}}
\end{center}
% ]
\appendix

\section{Surrogates Implementation Details}
\label{sur_get_appen}

\subsection{GAN Implementation}
For the training of the GANs we have used conditional generative network, with three hidden layers of size 100 and conditional discriminative network with two hidden layers of size 100. For all the hidden layers except the last one we have used $tanh$ activation. For the last hidden layer $leaky\_relu$ was used. The conditioning is performed via concatenating the input noise $\boldsymbol{z}$ with input parameters $\boldsymbol{\psi}$. The learning rate and batch size is set to $0.0008$ and $512$ correspondingly. We have used the idea from the \cite{gan_batch_size} to adjust the learning rate and a batch size for optimal training speed and performance. We have used Adam optimizer for both discriminator and generator with $\beta_1 = 0.5, \beta_2 = 0.999$. We have trained GAN only for 15 epochs for all the experiments. The number of discriminator updates per one generator update is $n_d=5$. In case of the Cramer GAN \cite{cramer_gan} we have used gradient penalty with $\lambda = 10$, discriminator output size equal to 256 and the number of discriminator updates $n_d$ is set to 1.

\subsection{FFJORD Implementation}

Training procedure and architecture of FFJORD model~\citep{ffjord_paper} were fixed for all experiments. We have used two hidden layers with 32 neurons each. The learning rate and batch size are set to $10^{-3}$ and 262144 respectively. It was trained with SWATS optimizer~\citep{Keskar2017ImprovingGP} until convergence, i.e. until log-likelihood is no longer improves for more than 200 epochs. For all hidden layers \textit{tanh} was used as nonlinearity, batch normalization lag were set to $10^{3}$ and \textit{fixed\_adam} were used as ODE solver. Usage of adaptive ODE solver and/or more elaborate choice of architecture probably could improve performance of the algorithm, but, firstly, it is out of the scope of our work, and, secondly, we were aiming to show that even without tuning for specific problem algorithm could shows performance comparable with recent works.

Original version of FFJORD does not have a support of conditional input. To address this issue we rewrote one of the base layers that were used in FFJORD library. We have added additional two-layers network with hidden dimensionality equal 8 that takes as an input conditional information and injects it in base layer output as an additive bias term.

\subsection{Monitoring quality of the surrogate}
\label{train_stats_monitor}
\begin{figure}[H]
\includegraphics[width=\linewidth]{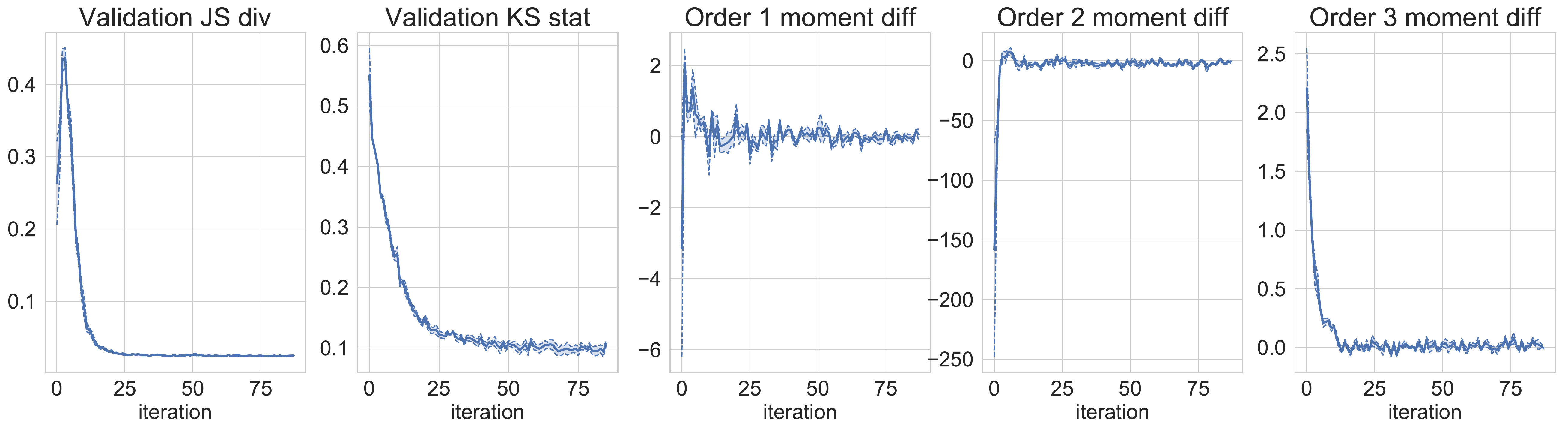}
\caption{An example of monitored statistics during surrogate training for one iteration of optimization. Left to right: Jensen–Shannon divergence, Kolmogorov–Smirnov statistic, difference between order one, two and three moment of the distributions from simulator and surrogate.}
\label{Fig:train_stats}
\end{figure}

During optimization we are constantly monitoring various statistics between samples from simulator and surrogate. The example of such statistics is presented in Fig~\ref{Fig:train_stats}. This is done to ensure that the surrogate learn a meaningful approximation of the simulator on each iteration of optimization and if this is not the case, the user can further tune the model.

\section{Bias of the estimator}
\label{bias_appendix}
The gradient bias is calculated per component of the gradient vector, i.e., if $\bpsi \in \mathbb{R}^ d$, then we present bias per component of this vector at each point of the optimization step. We calculate bias vector as follows:

\begin{minipage}{0.48\textwidth}
\begin{algorithm}[H]
\caption{Procedure to estimate the bias of the L-GSO}\label{alg:bisa_algo_comp}
\begin{algorithmic}[1]
\REQUIRE number N of $\bpsi$, number M of $\bx$ for surrogate training, number R of gradient estimates at point $\bpsi_t$, trust region $U_{\epsilon}$, size of the neighborhood $\epsilon$, Euclidean distance $d$
\FOR {t = 1, \dots, T}
    \STATE {$\mathbf{b} \leftarrow$ \O}
    \FOR {r = 1, \dots, R}
        \STATE \begin{varwidth}[t]{\linewidth}
        {Sample $\bpsi_i$ in the region $U_{\epsilon}^{\bpsi_t},\\
        i=1, \dots, N$}
        \end{varwidth}
        \STATE \begin{varwidth}[t]{\linewidth}
        \STATE {Sample training data $(\by^i_j, \bx^i_j, \bpsi_i),\\
        j=0, \dots, M$}
        \end{varwidth}
        \STATE \begin{varwidth}[t]{\linewidth}
        \STATE {Train generative model $S_{\theta} (\bz_l, \bx_l, \bpsi_l),\\
        \bz_l \sim \mathcal{N} (0,1), l = 1, \dots, MN $}
        \end{varwidth}
        \STATE \begin{varwidth}[t]{\linewidth}
        \STATE {Compute $\nabla_{\bpsi | \bpsi_t} \E[\mathcal{R}(\bar{\by})]$ from surrogate}
        \end{varwidth}
        \STATE {$\mathbf{b_r} \leftarrow \nabla_{\bpsi | \bpsi_t} \mathcal{R} (\by) - \nabla_{\bpsi | \bpsi_t}  \E[\mathcal{R}(\bar{\by})] $}
    \ENDFOR
\STATE {$\mathbf{bias}_t = \frac{1}{R} \sum_{r=1}^{R} \mathbf{b}_r$}
\STATE {$\mathbf{variance}_t = \frac{1}{R-1} \sum_{r=1}^{R} (\mathbf{b}_r - \mathbf{bias}_t)^2$}
\ENDFOR
\end{algorithmic}
\end{algorithm}
\end{minipage}
\begin{minipage}{0.48\textwidth}
\begin{figure}[H]
\centering
\includegraphics[width=0.8\linewidth]{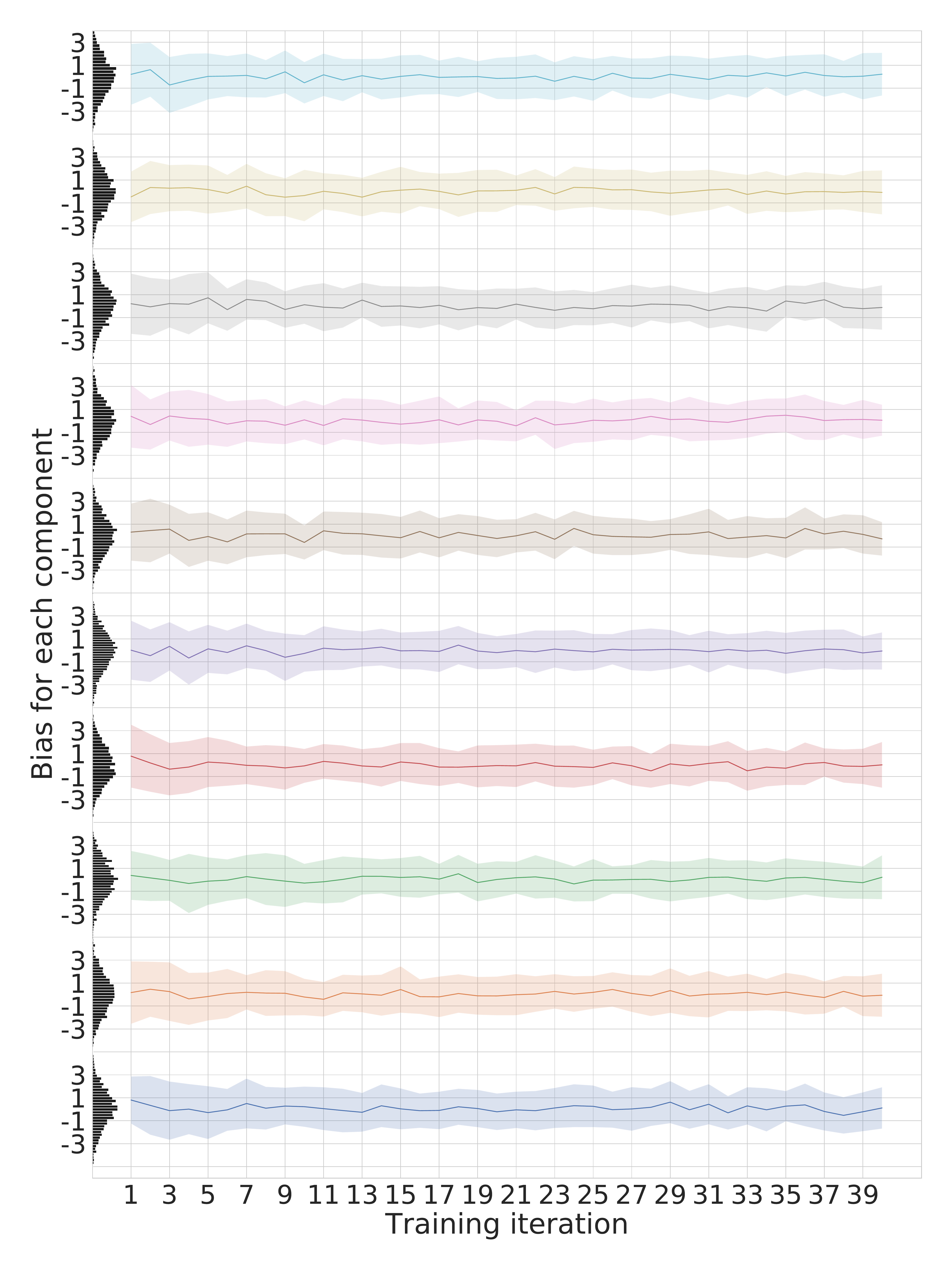}
\caption{The bias (solid line) and one standard deviation (shaded region) of the GAN based L-GSO gradient estimate for all dimensions of $\bpsi$ of the 10D Rosenbrock problem is shown as a function of the training step. The gray histograms shows the empirical distribution of bias averaged over all training iterations.}
\label{bias_path_all_components}
\end{figure}
\end{minipage}

The bias and variance for all parameters in the 10 dimensional Rosenbrock problem are presented in the Figure \ref{bias_path_all_components}.

\section{Optimization Implementation Details}
\label{appendix_opt_details}

Throughout all the experiments Adam~\citep{adam_opt} optimizer with default hyperparameters and learning rate equal $10^{-1}$ was used to perform update of the $\bpsi$ parameters.

Latin Hypercube sampling window of size $\epsilon=0.2$ was used for the ``Rosenbrock'', ``Submanifold Rosenbrock'', ``Nonlinear Submanifold Hump'', and ``Neural Network Weights Optimization'' problems, $\epsilon=0.5$ was used for ``Three Hump'' problem.

\subsection{Procedure For Mixing Matrix Generation}
\label{mixing_matrix_generation}
10-dimensional mixing matrix $A$ could be generated with the following Python code:
\begin{lstlisting}[language=Python,basicstyle=\small\ttfamily]
import numpy as np
def generate_orthogonal(in_dim, out_dim, seed=1337):
    assert in_dim > out_dim
    mixing_covar, _ = np.linalg.qr(np.random.randn(n_dim,out_dim))
    return mixing_covar
\end{lstlisting}

\subsection{Procedure For Initialization of Neural Network For Boston Regression Problem}

Neural network for Boston regression problem initialized as a two-layer network with $\tanh$-nonlinearity with predefined weights, using PyTorch.

\begin{lstlisting}[language=Python,breaklines,basicstyle=\small\ttfamily]
import torch
from torch import nn
def make_boston_net():
    torch.manual_seed(1337)
    net = nn.Sequential(
    nn.Linear(13, 6),
    nn.Tanh(),
    nn.Linear(6, 1)
    )
    initial_weights = torch.tensor(
        [0.0215, 0.0763, 0.0879, 0.0102, 
        0.095, 0.0508, 0.088, 0.101, 
        0.0782, 0.0684, 0.0658, 0.0509, 
        0.0207, 0.0618, 0.0756, 0.00784, 
        0.0968, 0.0685, 0.0113, 0.0745, 
        0.00154, 0.0772, 0.0472, 0.000906, 
        0.0723, 0.0779, 0.0594, 0.0785, 
        0.0918, 0.0634, 0.0853, 0.105, 
        0.00407, 0.0789, 0.0035, 0.0581, 
        0.0375, 0.0632, 0.0669, 0.00293, 
        0.0901, 0.0208, 0.0388, 0.0893, 
        0.00104, 0.0598, 0.0745, 0.08, 
        0.0283, 0.0106, 0.0371, 0.0667, 
        0.0331, 0.0356, 0.0661, 0.0554, 
        0.084, 0.0398, 0.00286, 0.0281, 
        0.0246, 0.0208, 0.0358, 0.033, 
        0.0421, 0.0505, 0.00544, 0.0269, 
        0.00527, 0.0569, 0.00538, 0.0786, 
        0.102, 0.0452, 0.0444, 0.105, 
        0.0765, 0.0689, 0.0249, 0.0933, 
        0.037, 0.0762, 0.0882, 0.0505, 
        0.0688, 0.0666, 0.101, 0.0857, 
        0.0488, 0.0303, 22.5328])
    net[0].weight.data = initial_weights[: 6 * 13].view(6, 13).detach().clone().float()
    net[0].bias.data = initial_weights[6 * 13: 6 * 13 + 6].view(6).detach().clone().float()

    net[2].weight.data = initial_weights[6 * 13 + 6: 6 * 13 + 6 + 6].view(1, 6).detach().clone().float()
    net[2].bias.data = initial_weights[6 * 13 + 6 + 6: 6 * 13 + 6 + 6 + 1].view(1).detach().clone().float()
    net.requires_grad_(False)
    return net
\end{lstlisting}

\subsection{Numerical Derivatives}

To obtain numerical derivatives of $\mathcal{R}$ we are using central difference scheme:

\begin{gather}
%\begin{gathered}
    f'_{\psi_i} \approx \left( \bar{\mathcal{R}}(\psi_1, \dots, \psi_i + \epsilon, \dots, \psi_p) - \right.  \left. \bar{\mathcal{R}}(\psi_1, \dots, \psi_i - \epsilon, \dots, \psi_p) \right) / 2 \epsilon,
%\end{gathered}
\end{gather}

Where, $\bar{\mathcal{R}} = \frac{1}{N} \sum\limits_{i=1}^{N} \mathcal{R}(F(z_i, x_i; \boldsymbol{\psi})) , \ \ x_i \sim p(\mathbf{X}), \ z_i \sim p(\mathbf{Z})$. For all experiments we set $\epsilon=0.1$. We can not use small $\epsilon$ due to the stochastic nature of $\bar{\mathcal{R}}$ (see appendix \ref{grid_search_results}, where we compare results with different values of $\epsilon$).

% \newpage
\section{Details of the Physics Problem}
\label{Physics_problem_formulation}

\begin{figure}
\begin{minipage}{0.48\linewidth}
\includegraphics[width=0.9\linewidth]{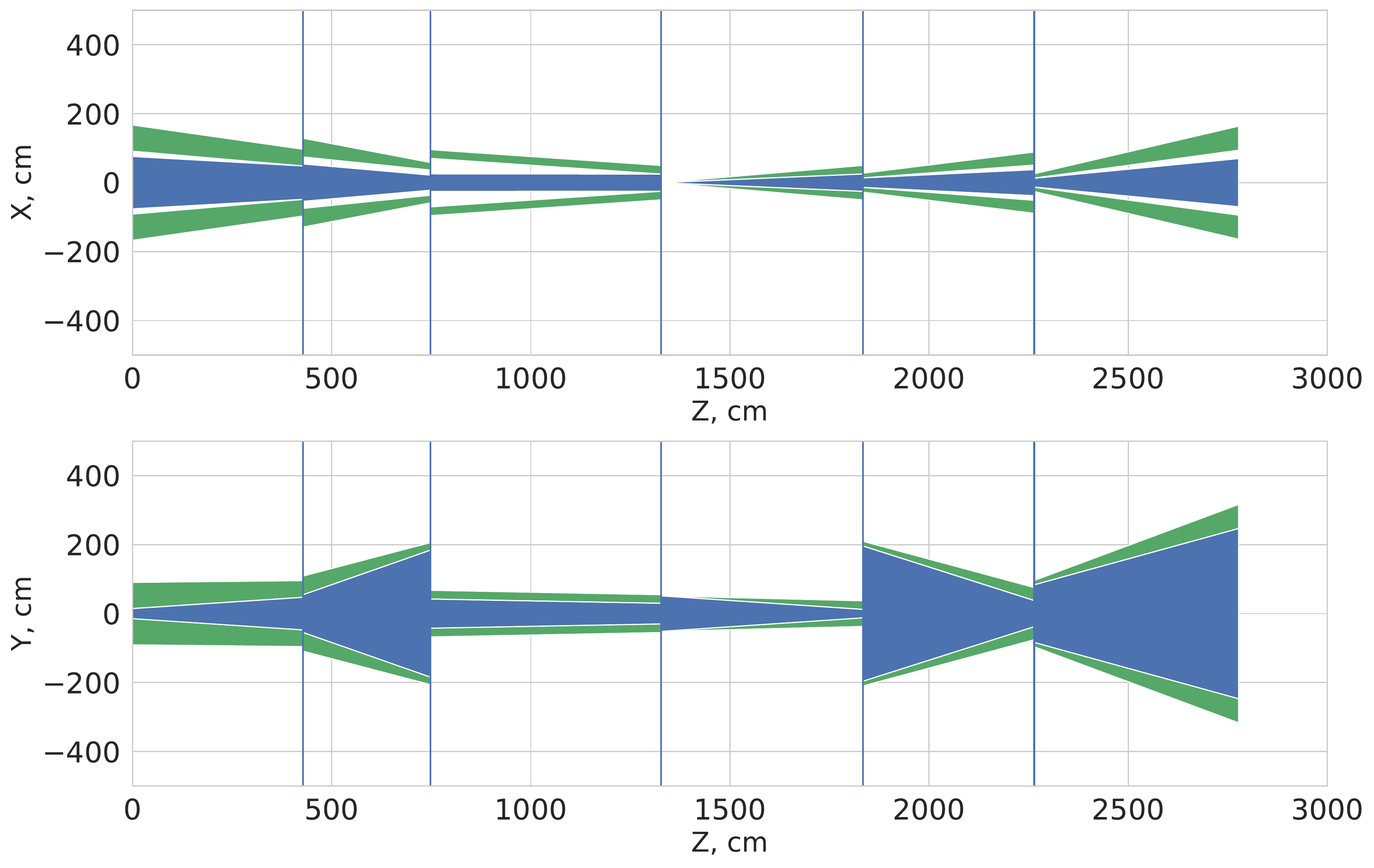}
\caption{The $x$-$z$ axes and $y$-$z$ axes profiles of the magnet system (the post optimization shape is shown). Animation of optimization process is available at \url{https://doi.org/10.6084/m9.figshare.11778684.v1}.}
\label{magnet_shape}
\end{minipage}
\hfill
\begin{minipage}{0.48\linewidth}
\begin{subfigure}[h]{0.5\linewidth}
\includegraphics[width=\linewidth]{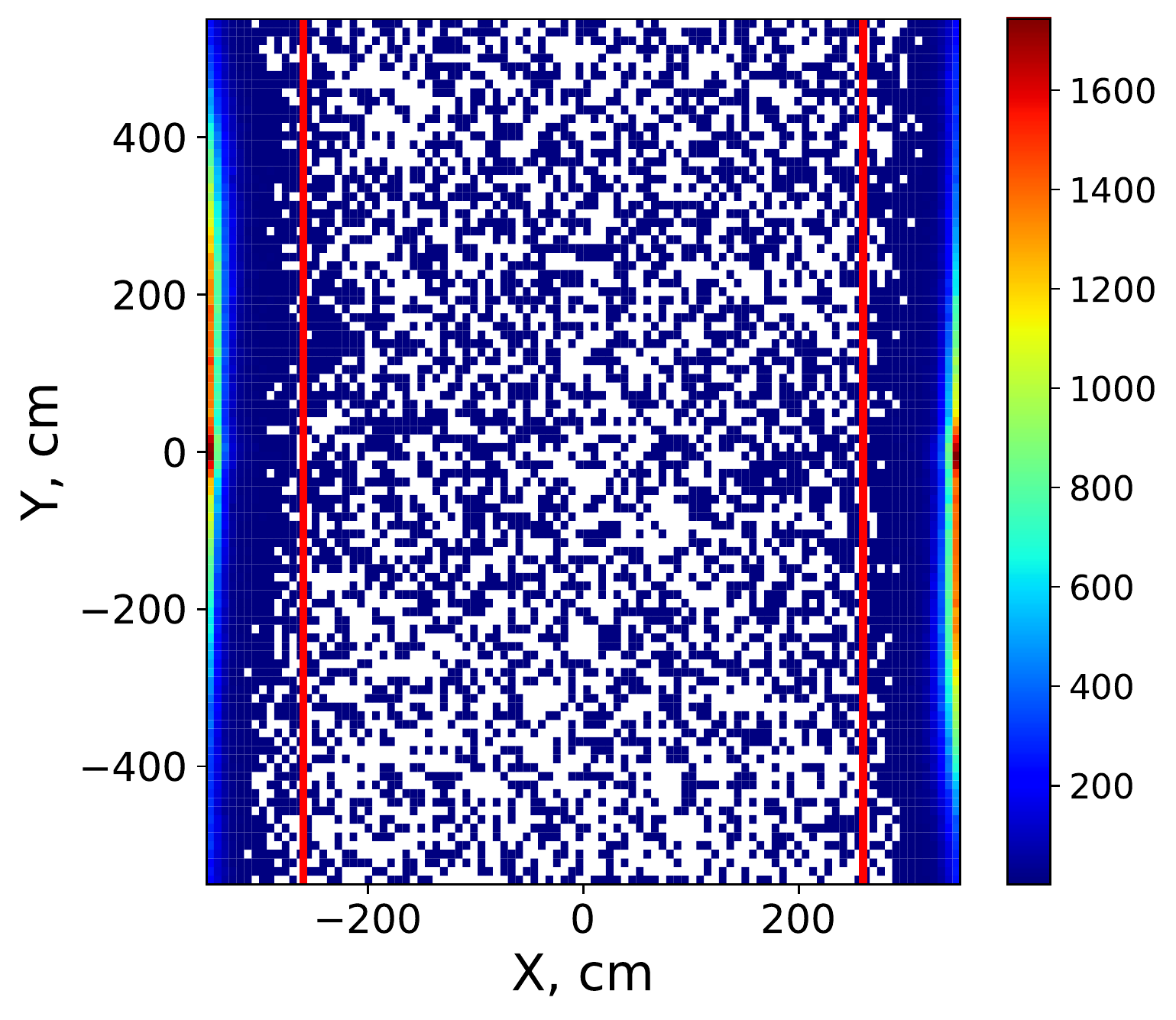}
\end{subfigure}
\hfill
\begin{subfigure}[h]{0.48\linewidth}
\includegraphics[width=\linewidth]{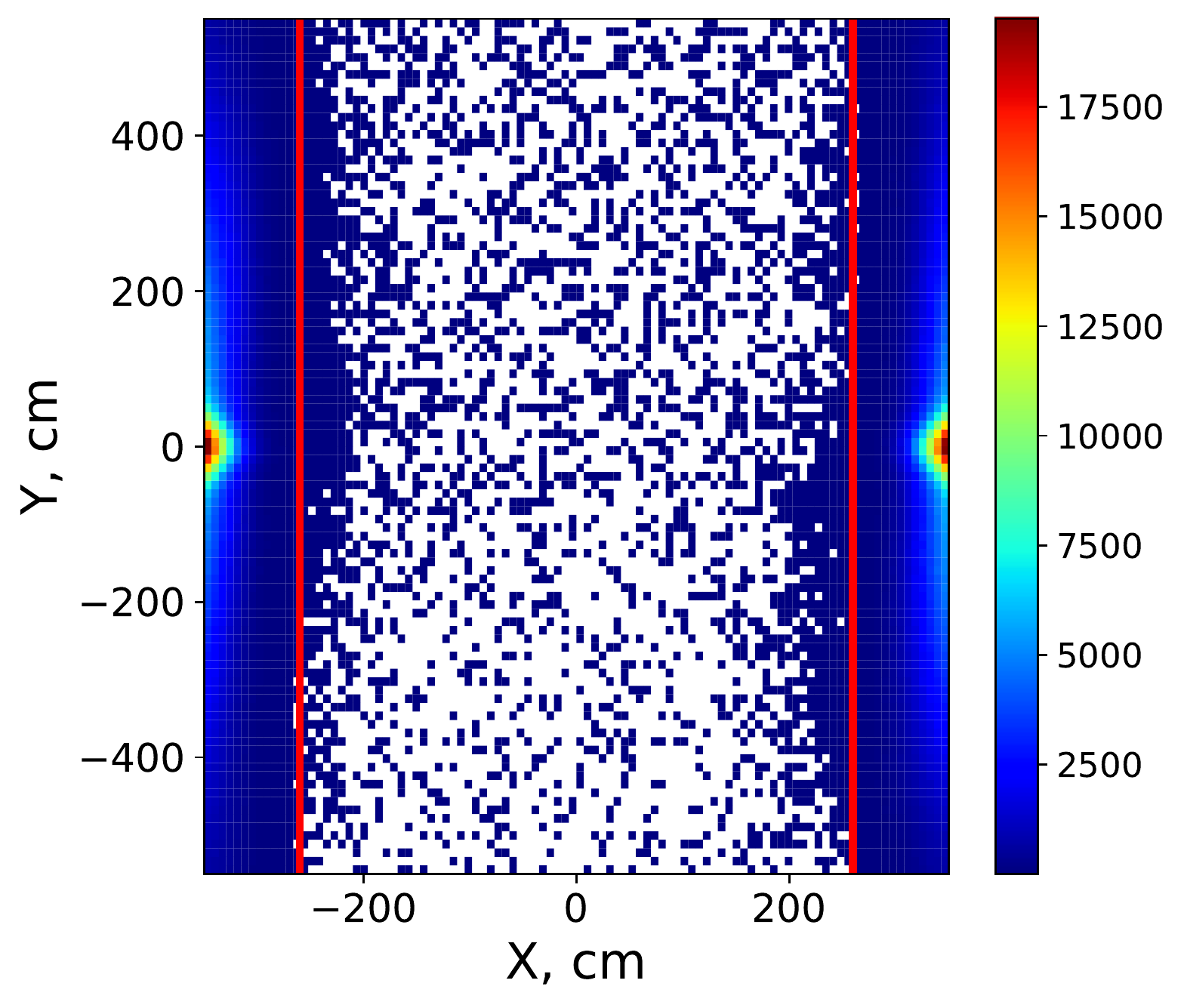}
\end{subfigure}%
\vspace{0.9cm}
\caption{Muon hits distribution in the detection apparatus (depicted as red contour) obtained by Bayesian optimization (Left) and by L-GSO (Right), showing better distribution. Color represents number of the hits in a bin.}
\label{hits_distr}
\end{minipage}%
\end{figure}

Muons are bent by the magnetic field and simultaneously experience stochastic scattering as they pass through the magnet which causes random variations in their trajectories. The coordinates perpendicular to incoming direction (the z-axis in Figure~\ref{magnet_shape}) are recorded. The magnet is constructed from six trapezoidal shapes with gaps each of which is described by seven parameters, as presented in Figure~\ref{magnet_shape}. Thus, for this task $\boldsymbol{\psi} \in \mathbb{R}^{42}$. Mathematically, formulation is $X = \{P, \phi, \theta, Q, C \}, X \in \mathbb{R}^7, \ \  y \sim \mathbb{R}^2$
where $y$ is a simulator output representing hit coordinates in the sensitive detector. X is sampled from empirical distribution $H$ (histogram), produced upfront. To make our optimization comparable with previously applied BO optimization, during optimization we have been working with subsample of $H$ of size of order of $O(500,000)$ events, same as in case of BO application. The objective function value reported in the Figure~\ref{ship_optimization} is calculated on this sample. To perform cross-validation of the obtained optima, we run physics simulation on the largest available sample, which does not contain samples from $H$. We have also validated the BO optima on the same available sample. The comparison is presented in Table~\ref{tab:ours_bo_ship}. Both BO and L-GSO have been compared on the simplified geometry of the experiment. The distributions of muons in the detection apparatus obtained by L-GSO is compared with BO optimization in Figure~\ref{hits_distr}. 

\begin{table}
\caption{Comparison of the optima, obtained by L-GSO and Bayesian optimization for the physics problem.}
\scriptsize
\centering
%\begin{tabularx}{\columnwidth}{cccc}
\begin{tabular}{cccc}
\toprule
Algorithm & Objective value & Magnet length (m) & Magnet weight (kt) \\ %\begin{tabular}[x]{@{}c@{}}Magnet\\weight (kt)\end{tabular} \\
\midrule
L-GSO & $\sim 2200$ & 33.39 & 1.05 \\
BO & $\sim 3000$ & 35.44 & 1.27 \\
\bottomrule
\end{tabular}
\label{tab:ours_bo_ship}
\end{table}

\section{Grid Search of Optimal Parameters}
\label{grid_search_results}

\subsection{Grid Search Hyperparameters For L-GSO}
\begin{figure}[ht]
\begin{subfigure}[h]{0.49\linewidth}
\includegraphics[width=\linewidth]{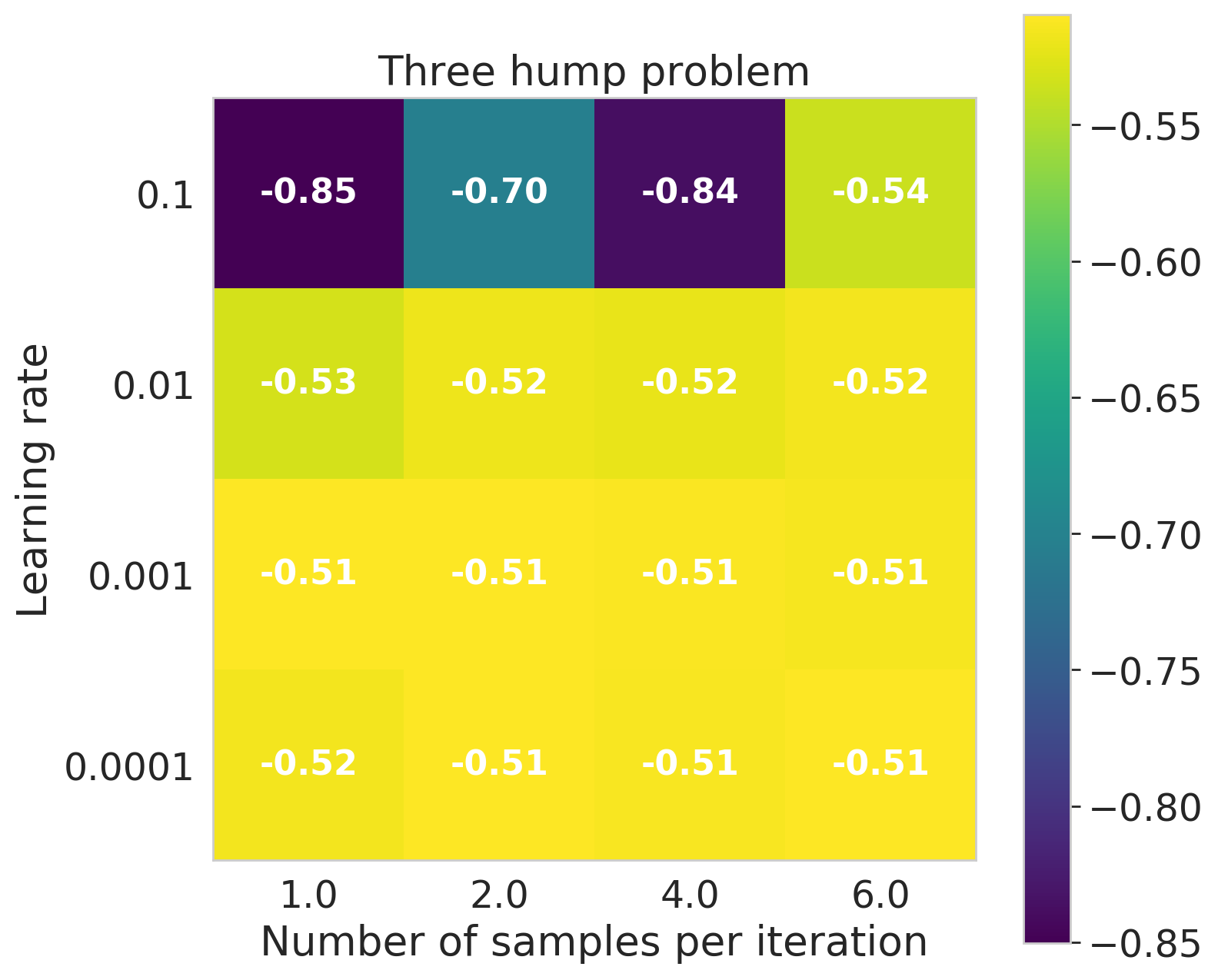}
\caption{}
\label{grid_search_hump}
\end{subfigure}
\hfill
\begin{subfigure}[h]{0.49\linewidth}
\includegraphics[width=\linewidth]{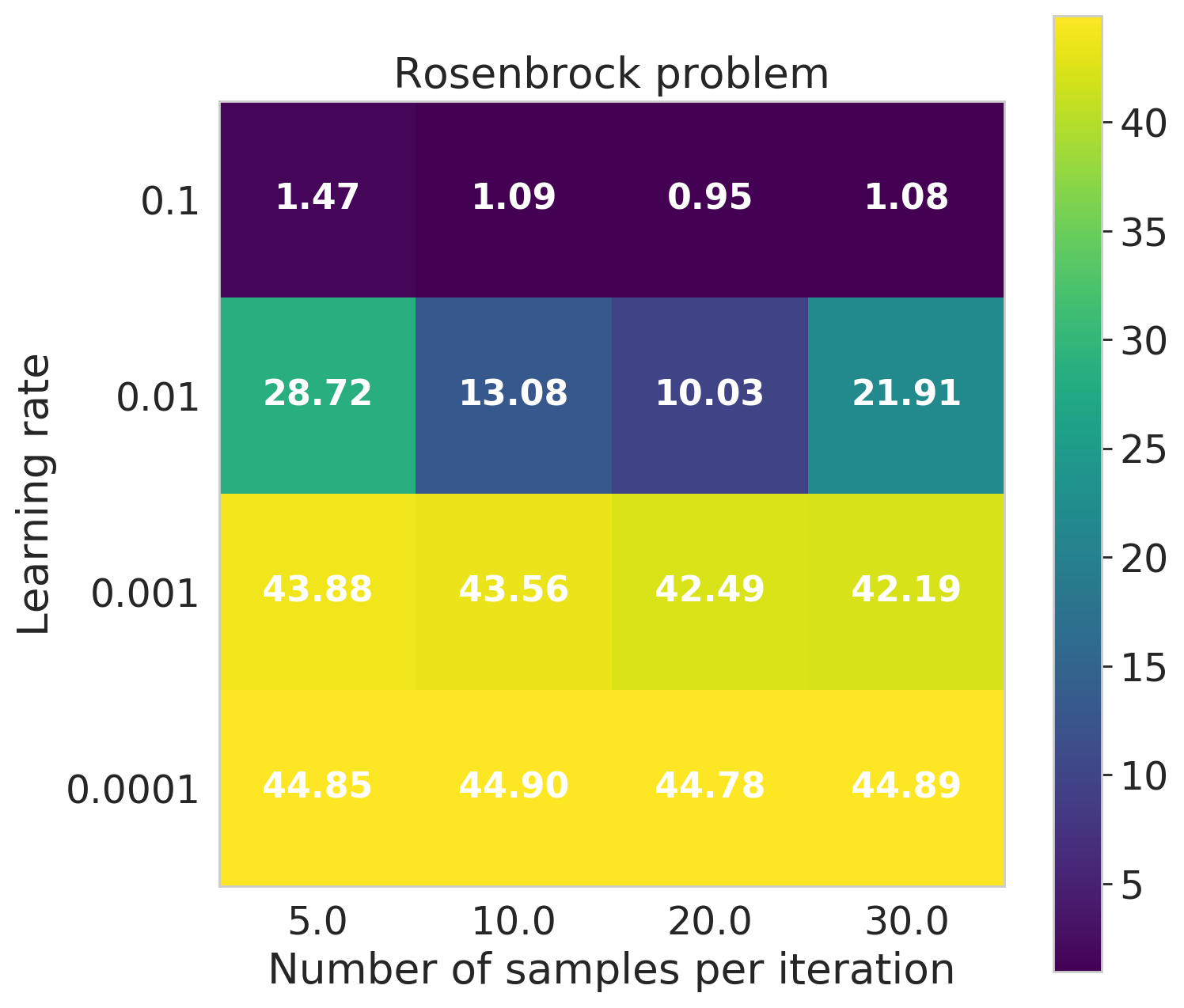}
\caption{}
\label{grid_search_rosenbrock}
\end{subfigure}%
\caption{Grid search of learning rate and number of samples for L-GSO. Color represents final quality for Three hump problem (Left) and for Rosenbrock problem (Right).}
\end{figure}

\begin{figure}[ht]
\begin{subfigure}[h]{0.49\linewidth}
\includegraphics[width=\linewidth]{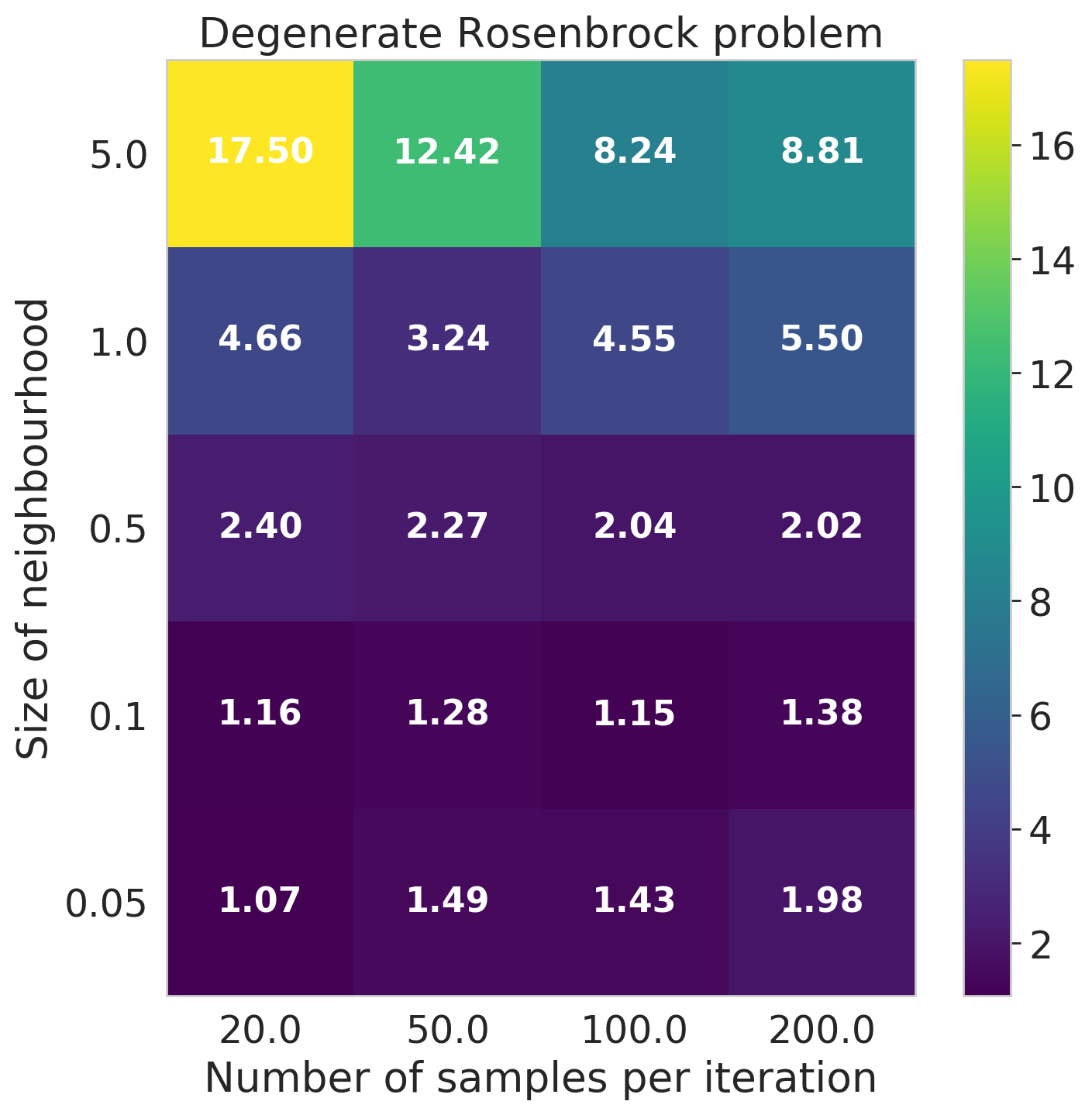}
\caption{Final value of objective function $\mathcal{R}$ of L-GSO for 100 dimensional Degenerate Rosenbrock problem.}
\label{rmd_gs_risk}
\end{subfigure}
\begin{subfigure}[h]{0.49\linewidth}
\includegraphics[width=\linewidth]{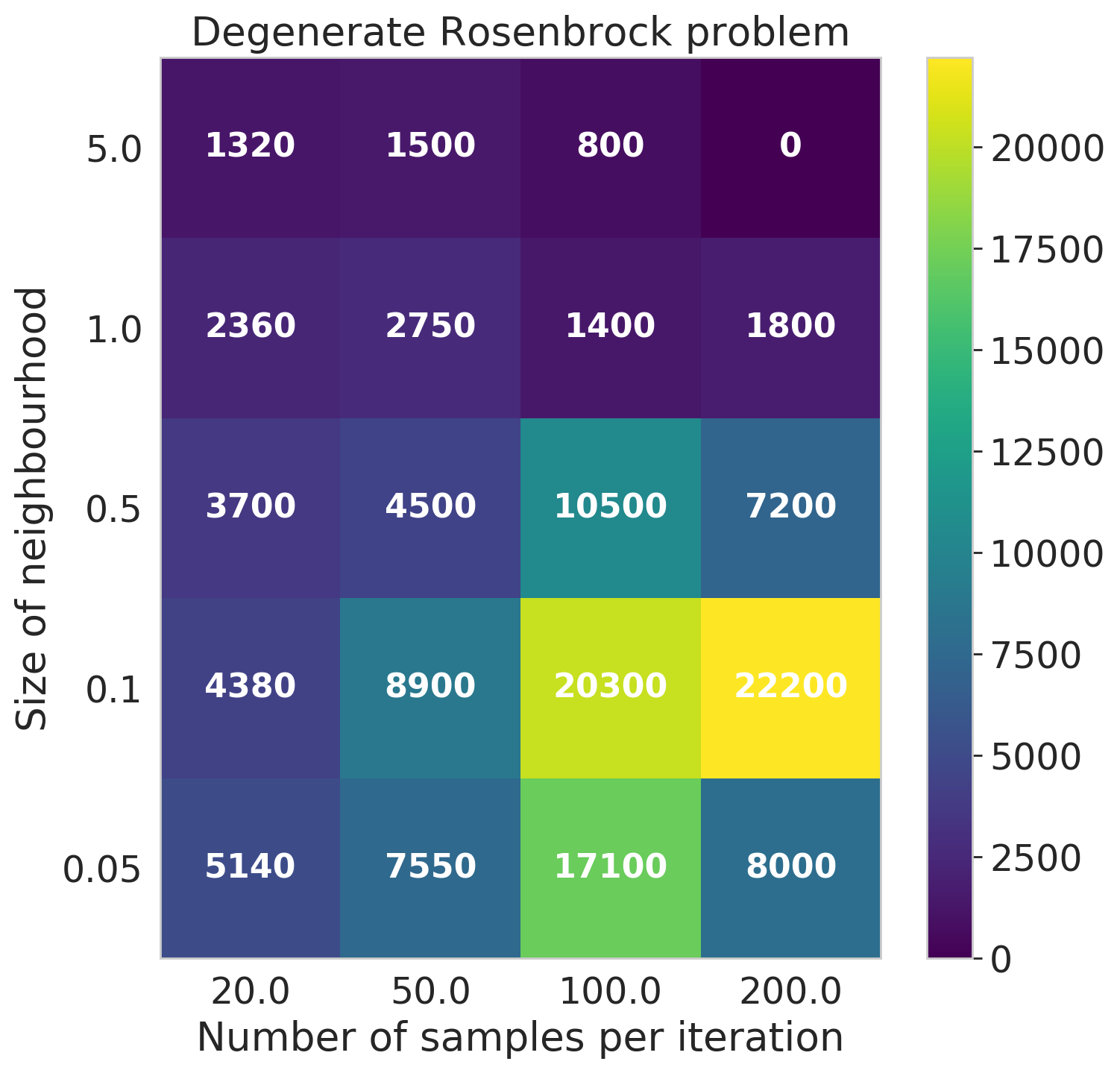}
\caption{Number of samples (calls of the simulator) needed by L-GSO to converge to final value of objective funciton.}
\label{rmd_gs_speed}
\end{subfigure}
\caption{}
\end{figure}

We have optimized crucial hyperparameters of L-GSO, such as learning rate, size of the sampling neighborhood $\epsilon$ and the number of samples of $\bpsi$ in this neighborhood with grid search. The grid search for Three hump and Rosenbrock problem is presented in Figure \ref{grid_search_hump}, \ref{grid_search_rosenbrock}. As it can be seen, for both problems best quality is obtained when number of samples is approximately equal to the dimensionality of the problem and when learning rate is close to 0.1.
We found that learning rate 0.1 is optimal for all the toy problems under consideration. Thus, we have fixed it to be 0.1 for other grid search experiments. In the Figure~\ref{rmd_gs_risk},~\ref{rmd_gs_speed} we present the grid search for 100 dimensional Degenerate Rosenbrock problem for number of samples per iteration and size of the neighborhood. We found that L-GSO is very sensitive to the size of the neighborhood $\epsilon$, whereas surprisingly robust to the number of samples, as it is seen in Figure~\ref{rmd_gs_risk}.

\subsection{Grid Search Hyperparameters For Numerical Differentiation}

\begin{figure*}
\begin{subfigure}[h]{0.49\linewidth}
\includegraphics[width=\linewidth]{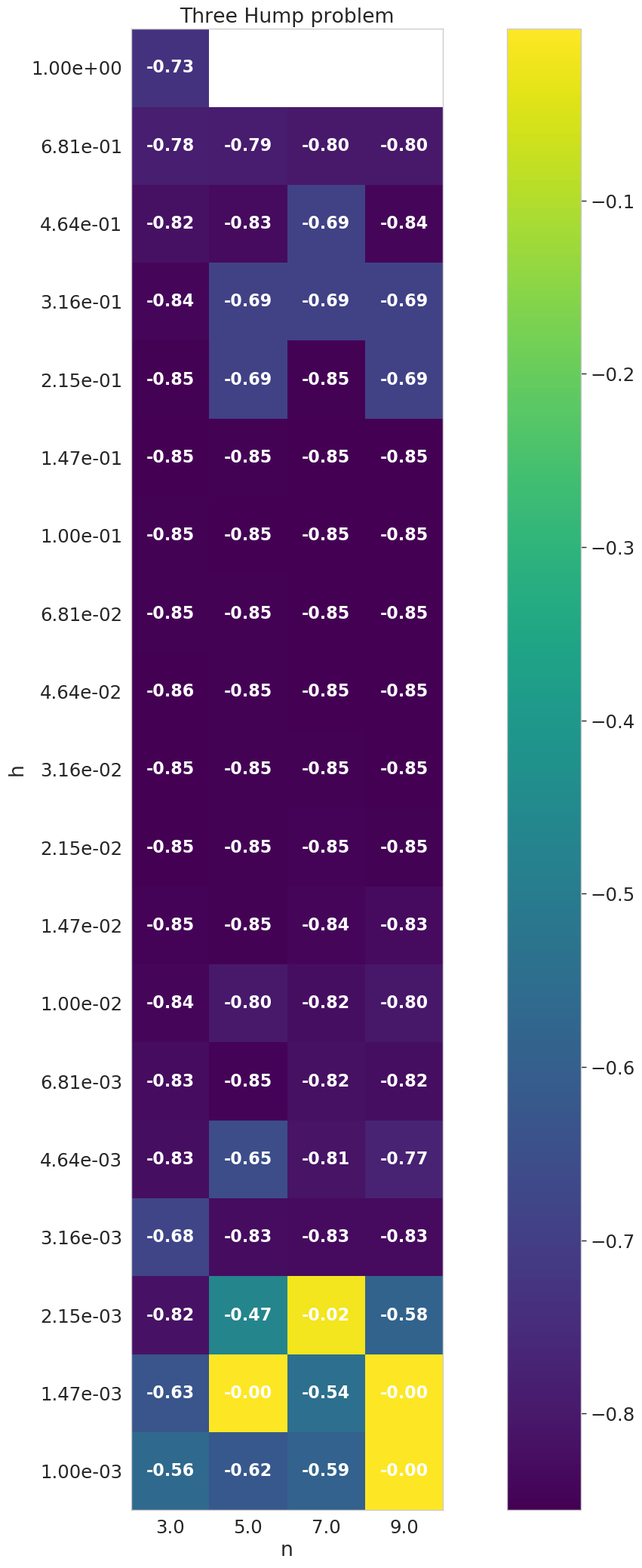}
\caption{Final value if objective function $\mathcal{R}$ of numerical differentiation for hump problem.}
\label{hump_gs_num_risk}
\end{subfigure}\quad
\begin{subfigure}[h]{0.49\linewidth}
\includegraphics[width=\linewidth]{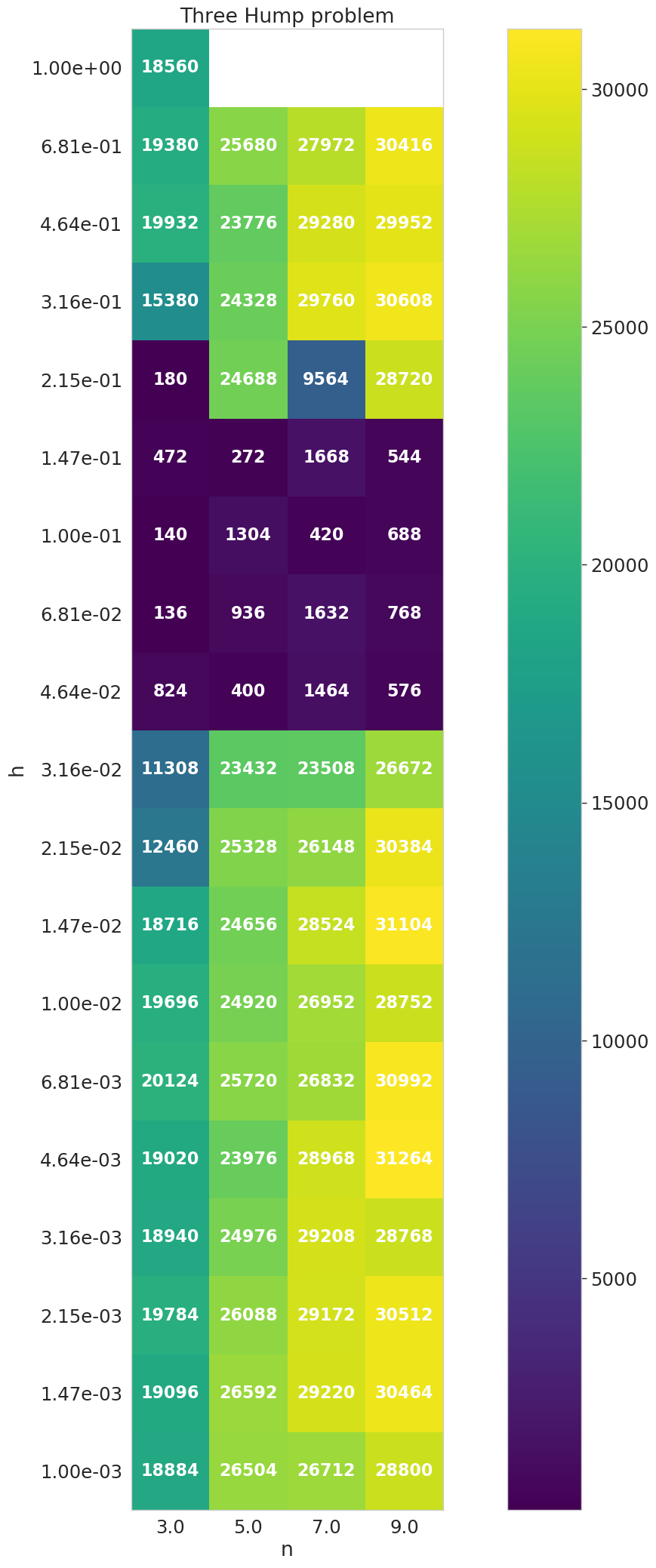}
\caption{Number of samples(calls to the simulator) needed by numerical differences to converge.}
\label{hump_gs_num_speed}
\end{subfigure}
\caption{}
\end{figure*}

\begin{figure*}
\begin{subfigure}[h]{0.47\linewidth}
\includegraphics[width=\linewidth]{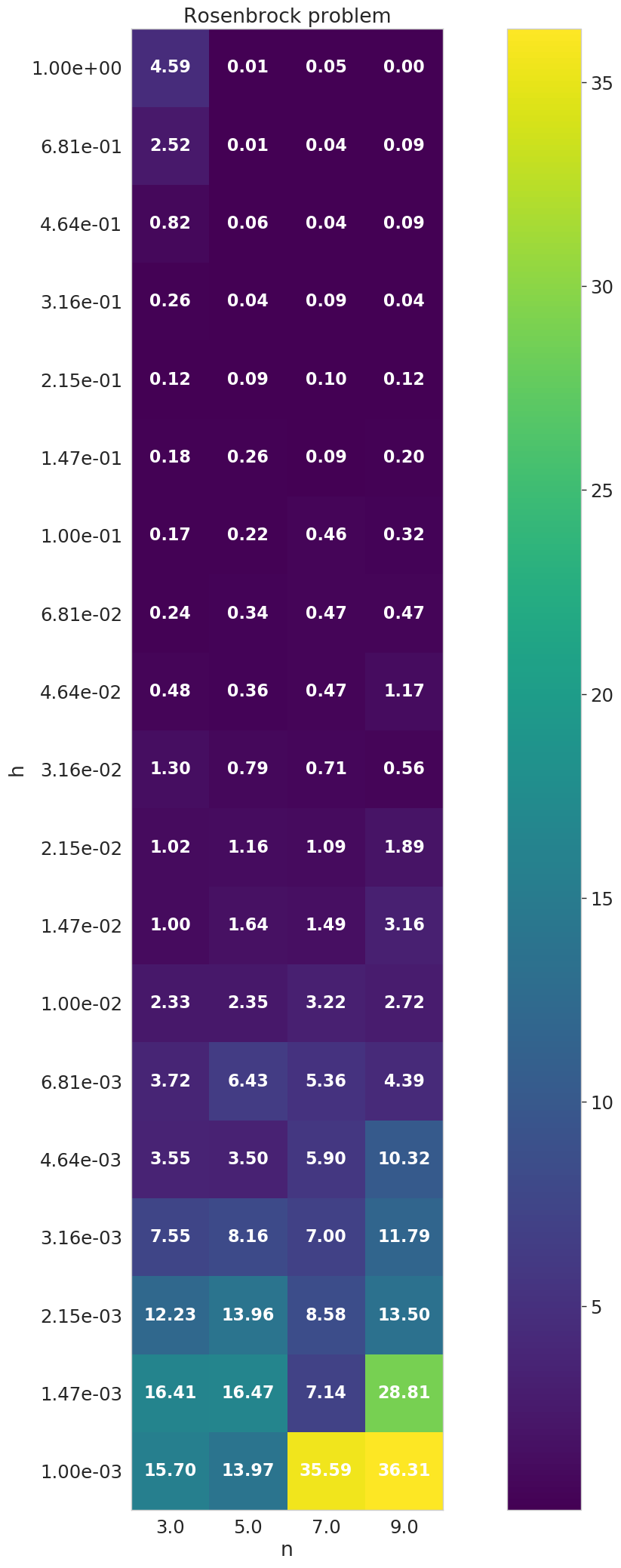}
\caption{Final value if objective function $\mathcal{R}$ of numerical differentiation for 10-dim Rosenbrock problem.}
\label{r_gs_num_risk}
\end{subfigure}\quad
\begin{subfigure}[h]{0.49\linewidth}
\includegraphics[width=\linewidth]{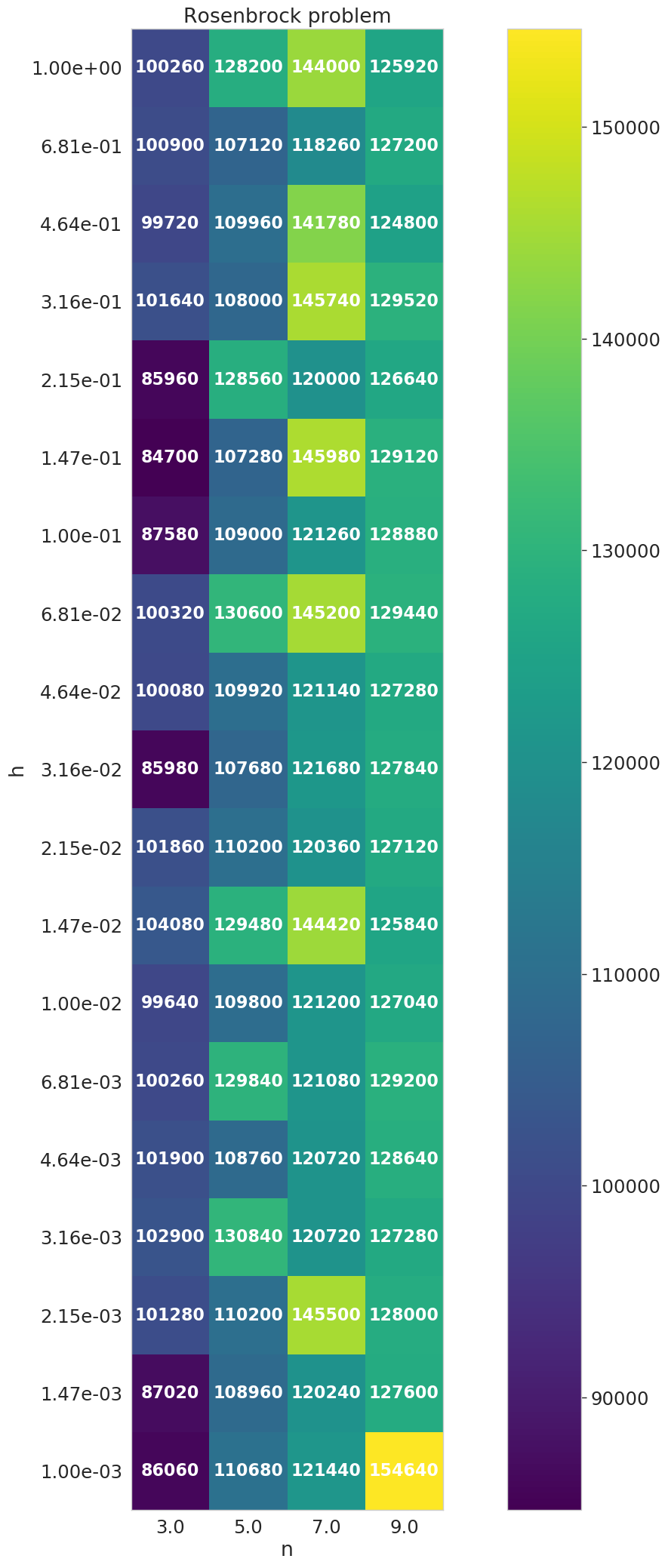}
\caption{Number of samples(calls of the simulator) needed by numerical differentiation to converge.}
\label{r_gs_num_speed}
\end{subfigure}
\caption{}
\end{figure*}

\begin{figure*}
\begin{subfigure}[h]{0.46\linewidth}
\includegraphics[width=\linewidth]{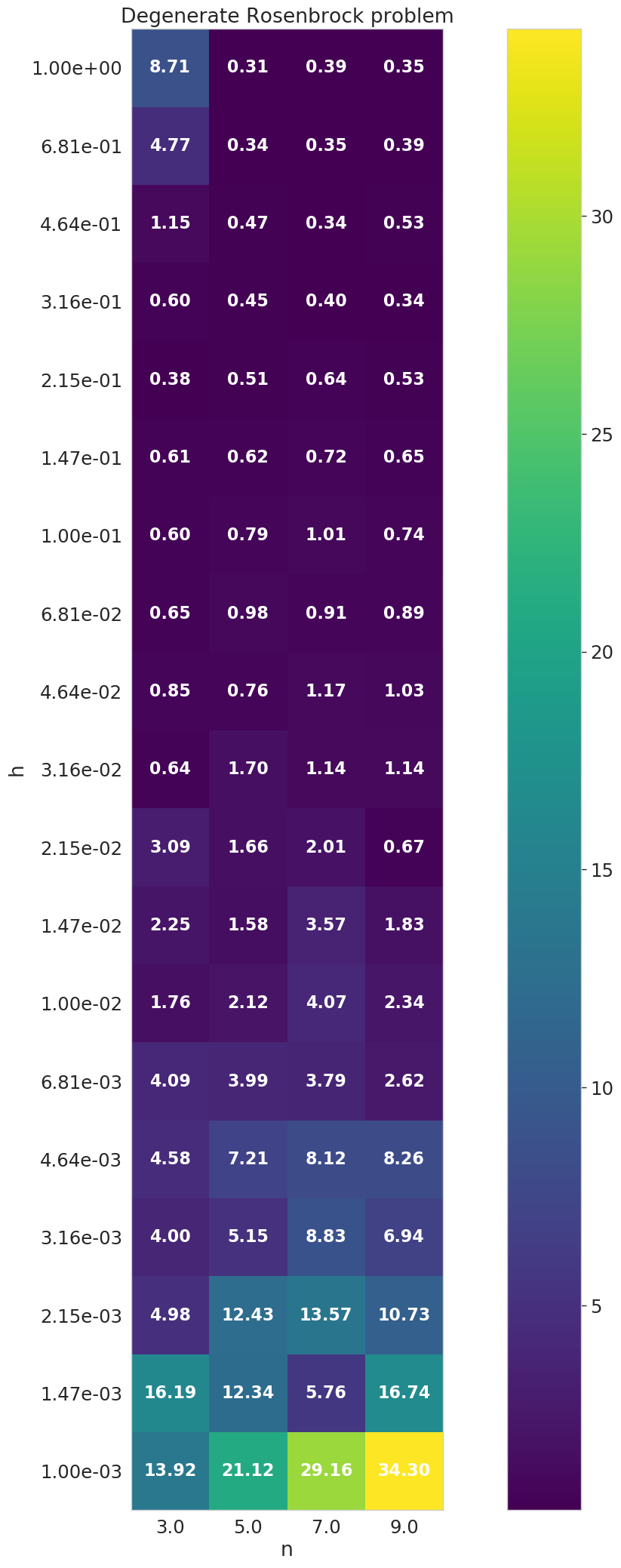}
\caption{Final value if objective function $\mathcal{R}$ of numerical differentiation for 100-dim Degenerate Rosenbrock problem.}
\label{rmd_gs_num_risk}
\end{subfigure}\quad
\begin{subfigure}[h]{0.49\linewidth}
\includegraphics[width=\linewidth]{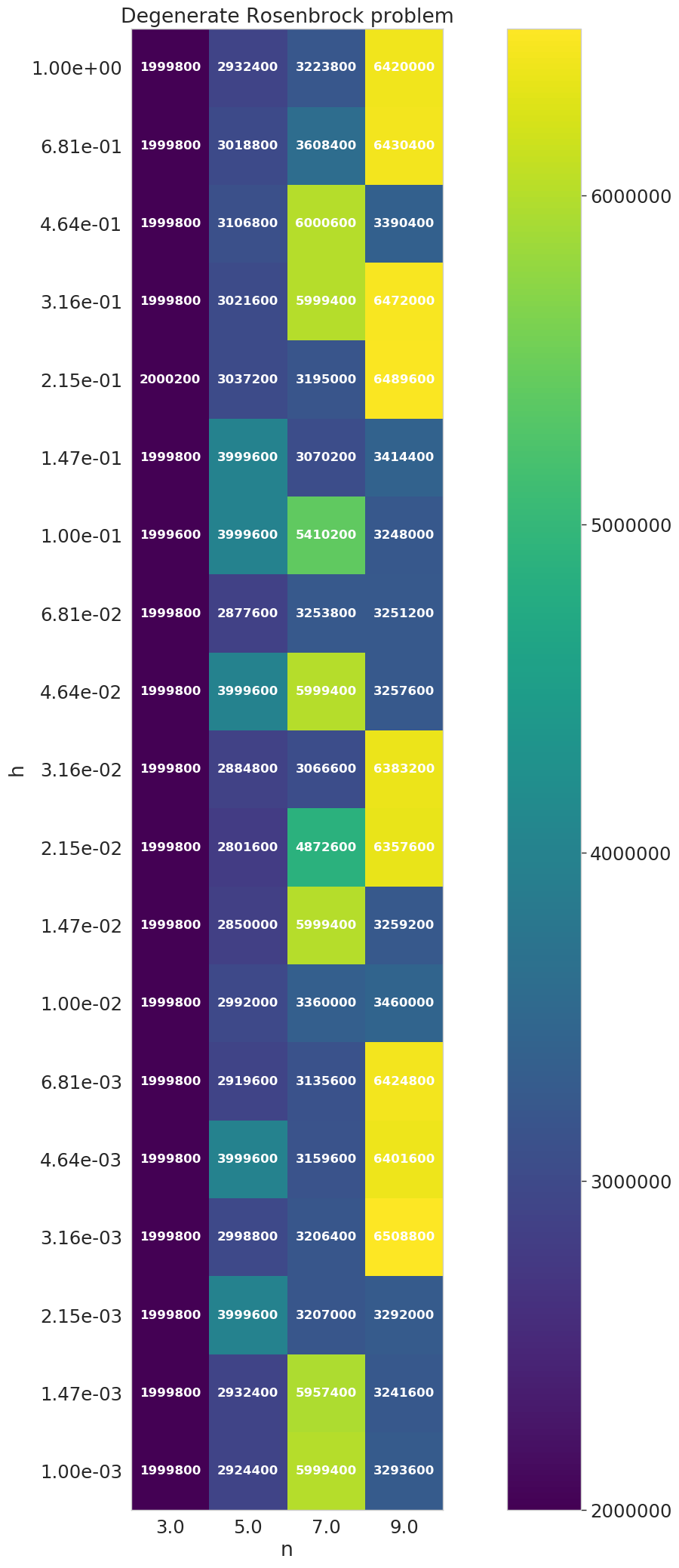}
\caption{Number of samples(calls of the simulator) needed by numerical differentiation to converge.}
\label{rmd_gs_num_speed}
\end{subfigure}
\caption{}
\end{figure*}

We performed grid search over the order of numerical scheme $n$ and step size $h$ for numerical optimization for all four toy problems. As an example, the results for the toys problems are presented in Figure ~\ref{hump_gs_num_risk},~\ref{hump_gs_num_speed}, %~\ref{lts_gs_num_risk}, ~\ref{lts_gs_num_speed},
~\ref{r_gs_num_risk},~\ref{r_gs_num_speed},
~\ref{rmd_gs_num_risk},~\ref{rmd_gs_num_speed}.

\end{document}